%% file: main.tex
\documentclass[10pt,twocolumn,letterpaper]{article}

\usepackage[pagenumbers]{cvpr}      

\definecolor{cvprblue}{rgb}{0.21,0.49,0.74}
\usepackage[pagebackref,breaklinks,colorlinks,allcolors=cvprblue]{hyperref}

\usepackage{amssymb}
\usepackage{amsmath}
\usepackage{natbib}
\usepackage{listings}
\usepackage{booktabs}
\usepackage{float}
\usepackage{tabularx}
\usepackage{pifont}
\usepackage{xcolor}
\usepackage[most]{tcolorbox}
\tcbuselibrary{skins}
\usepackage{algorithm}
\usepackage{algpseudocode}
\usepackage{xcolor}
\usepackage{url}
\usepackage{array}
\newcolumntype{C}{>{\centering\arraybackslash}X}
\usepackage{multirow}
\usepackage{tabularx}
\usepackage{amssymb}
\usepackage{bm}
\usepackage{array}
\usepackage[table]{xcolor}

\definecolor{handlecolor}{RGB}{200, 30, 30}    
\definecolor{blockcolor}{RGB}{30, 80, 180}     
\definecolor{commentcolor}{RGB}{120, 120, 120}

\newcommand{\block}[1]{\textcolor{blockcolor}{\textbf{#1}}}

\def\paperID{*****} 
\def\confName{CVPR}
\def\confYear{2026}

\title{SAGE: An LLM-driven Self Reflective Agentic Framework for Fraud Detection}

\author{Yichen Chen\\
National University of Singapore\\
School of Computing, Singapore 117417\\
{\tt\small chenyichen@u.nus.edu}
\and
Siying Li\\
University of Chinese Academy of Sciences\\
Institute of Information Engineering, Beijing 100085\\
{\tt\small lisiying@iie.ac.cn}
\and
Yuhang Liang\\
China Mobile Communications Group\\
Big Data Business Group, Beijing 102206\\
{\tt\small liangyuhang@chinamobile.com}
\and
Lijun Wang\\
China Mobile Communications Group\\
Big Data Business Group, Beijing 102206\\
{\tt\small wanglijunit@chinamobile.com}
\and
Renyang Liu\thanks{Corresponding author.}\\
National University of Singapore\\
Institute of Data Science, Singapore 117602\\
{\tt\small ryliu@nus.edu.sg}
}

\begin{document}
\maketitle
\input{sections/00_abstract}
\input{sections/01_introduction}
\input{sections/02_background}
\input{sections/03_preliminaries}
\input{sections/04_methodology}
\input{sections/05_experiments}
\input{sections/06_discussion}
\input{sections/07_conclusion}
{
    \small
    \bibliographystyle{ieeenat_fullname}
    \bibliography{refs}
}


\end{document}

%% file: sections/00_abstract.tex
\begin{abstract}
Fraud detection in payment, e-commerce, and telecommunications systems requires accuracy at the individual level, robustness under severe class imbalance, and ease of understanding for risk managers. Existing methods fall at least one of these requirements: automated machine learning systems search a fixed numerical space without semantic awareness of the dataset; graph neural network-based methods require pre-defined relational graphs and remain opaque at the individual-decision level; and the design of general-purpose large language model (LLM) agents does not consider the recall and precision constraints specific to real-world fraud detection. In this paper, we propose SAGE, the first end-to-end LLM-driven multi-agent framework for fraud detection. SAGE coordinates three dedicated agents that make decisions based on a six-layer Data Diagnostic Tree (DDT) and a Markov decision process guided by natural-language gradients, automatically optimizing the model under a fraud-specific reward. On five fraud datasets and five LLM backbones, SAGE wins $96.00\%$ of method--dataset comparisons and improves F1 by an average of $40.86\%$ over baselines. The code is available at \url{https://github.com/yichenC1c/SAGE}.
\end{abstract}
\vspace{-8pt}

%% file: sections/01_introduction.tex
\section{Introduction}
\label{sec:introduction}

Fraud has become one of the most prevalent economic threats in the digital age, severely eroding public trust in payment systems, e-commerce platforms, and telecommunications networks. The rapid development of large language models (LLMs) and increasingly automated fraud strategies have led to an exponential increase in the operational costs of fraud cases. Therefore, fraud detection needs to shift from back-end data mining tasks to automated engineering processes that are responsive and sustainable, while maintaining interpretability for risk managers and auditors. However, in most real-world environments, the available evidence for each suspicious entity is not a rich relational graph, but rather structured features such as device fingerprints, transaction history, call patterns, and aggregated behavioral statistics. The core problem this paper aims to address is: how to automatically build accurate, robust, and interpretable fraud detectors based on such individual-level data.

While numerous studies have dedicated themselves to addressing this issue, existing research still falls short. Traditional supervised processes and automated machine learning (AutoML) systems excel on structured data, but they rely on fixed numerical searches, and the generated models lack a semantic foundation. Some anti-fraud methods based on graph neural networks (GNNs) require pre-defined relationship graphs, which are costly to construct in practice and remain opaque at the individual decision-making level, making them difficult to meet the needs of real-world anti-fraud operations. Recently developed general-purpose LLM agents can plan, encode, and self-correct, but they are all developed based on general benchmarks, rather than on the recall and precision-constrained task environments required for anti-fraud. To our knowledge, there is currently no AI agent framework specifically designed for classifier construction tasks in real-world anti-fraud scenarios.

This gap prompted us to propose an innovative design concept: building an LLM agent specifically for constructing fraud detection classifiers for real-world business scenarios. In this agent, the model has sufficient autonomy to reason about data features and the classifier itself, while being subject to structural and reward-based constraints, ensuring its decisions are evidence-based and consistent with anti-fraud requirements. Implementing this design concept faces two specific challenges:
\begin{itemize}\setlength\itemsep{3pt}
    \item \textbf{C1:} How can the agent quickly construct a global, realistic picture of the fraud dataset without being overwhelmed by the massive amount of raw column-level data, thus selecting an efficient and correct classifier model?
    \item \textbf{C2:} How can the agent's iterative process adhere entirely to principles rather than being arbitrary, ensuring that each optimization is based on the task objective and existing evidence?
\end{itemize}

To answer these questions, we propose SAGE, an end-to-end multi-agent framework for structured data fraud detection. SAGE decomposes the workflow into three dedicated agents driven by LLM, operating strictly in sequence: first, the ``Profiling Agent'' interprets the dataset into a Data Diagnostic Tree (DDT) to semantically perceive the full features of the fraud dataset, avoiding agent context loss due to excessively long datasets by profiling the dataset; second, the ``Planning Agent'' uses the DDT described by the ``Profiling Agent'' to understand the overall dataset, selects the optimal algorithm, and synthesizes an initial classifier model customized for the dataset; finally, the ``Optimization Agent'' iteratively optimizes the model in the code space through a finite-time Markov decision process. In this process, the language model first emits a natural language gradient to evaluate the current model, then transforms this evaluation into localized code optimization, ultimately achieving the recall and precision constraints required for the fraud business scenario set by the reward mechanism. Users only need to provide the fraud dataset; the entire process, from data analysis to model optimization, can be completed without human intervention, saving anti-fraud experts a significant amount of time previously spent on data cleaning, feature engineering, and parameter tuning.

In summary, our main contributions are as follows:
\begin{itemize}\setlength\itemsep{3pt}
    \item As far as we kown, SAGE is the first end-to-end LLM-based multi-agent framework designed specifically for the detection of individual-level tabular fraud datasets.
    \item We introduce the DDT, a six-layer structured prior that characterizes the feature information of extremely long raw datasets as a fraud-aware tree-structured representation, allowing the agent to make dataset-grounded decisions within a limited context window.
    \item We formalize agent code optimization as a finite-horizon MDP driven by natural-language gradients, in which natural language is translated into concrete code modifications under fraud-specific recall and precision constraints.
    \item We evaluate SAGE on five fraud datasets (four public benchmarks and one real-world industrial dataset) and across five LLM backbones. The results show that it significantly outperforms existing AutoML systems, LLM coding agents, and human experts, while remaining insensitive to changes in the underlying language model.
\end{itemize}

The remainder of this paper is organized as follows. Section~\ref{sec:back_ground} analyzes the current background and existing fraud detection paradigms, and compares related works. Section~\ref{sec:preliminaries} formalizes the anti-fraud problem and introduces the modeling primitives that run through SAGE. Section~\ref{sec:methodology} details the SAGE framework and its design principles, including the Data Diagnostic Tree and the natural-language-gradient-guided Markov decision process (MDP). Section~\ref{sec:experiments} introduces the experimental setup, main results, robustness and sensitivity analysis, interpretability case study, and ablation study. Section~\ref{sec:discussion} discusses the limitations and future research directions, and Section~\ref{sec:conclusion} summarizes the entire paper.

%% file: sections/02_background.tex
\section{Related Work}
\label{sec:back_ground}

\subsection{Fraud detection background}
\label{sec:rw_fraud}
Fraud has become one of the most destructive economic threats worldwide, permeating both financial and telecommunications sectors. A 2025 survey of 46{,}000 adults across 42 markets found that 57\% of respondents had experienced fraud in the preceding year, with estimated global losses reaching \$442 billion~\cite{gasa2025scams}. Fraud takes many forms: credit-card and payment fraud, e-commerce fraud, account takeover, cryptocurrency transaction fraud, and, in telecommunications, voice phishing and SMS phishing (smishing). A substantial body of work has formalized these settings: \cite{dalpozzolo2018realistic} formalizes the credit-card fraud detection system; \cite{xiang2023semisupervised} addresses credit-card fraud through a semi-supervised gated attention network over transaction graphs; \cite{xiao2024vecaug} targets camouflaged frauds in tabular data through cohort augmentation; \cite{he2026factguard} introduces an event-centric, commonsense-guided framework for fake detection; and \cite{telecomllm2026} targets covert textual expressions of telecom fraud. Across these settings, the detection problem reduces to individual-level classification: judging whether a single transaction, account, or subscriber is fraudulent from its own behavioral record, typically encoded as device attributes, transaction records, calling behavior, and aggregated temporal/frequency statistics. This task carries a difficult statistical profile: fraud is extremely rare (legitimate activity outnumbers fraud by tens to hundreds to one), a severe imbalance recognized as a fundamental learning obstacle~\cite{he2009learning} and repeatedly identified as the central challenge of fraud detection~\cite{lucas2020survey, yousefi2019comprehensive}; the cost of errors is asymmetric under a strict false-alarm budget; and fraud is adversarial and non-stationary~\cite{dalpozzolo2014learned}. These properties are reflected in widely used benchmarks such as the European credit-card dataset~\cite{dal2015calibrating}, the PaySim simulator~\cite{lopezrojas2016paysim}, and the Elliptic Bitcoin dataset~\cite{weber2019antimoney}.

\subsection{Fraud detection methods}
\label{sec:rw_classical}

Classical research has demonstrated the power of data mining classifiers: \cite{bhattacharyya2011data} compared logistic regression, support vector machines, and random forests on real credit-card data, and \cite{xuan2018random} confirmed the effectiveness of random forests; sequence-aware variants such as the LSTM model of~\cite{jurgovsky2018sequence} and the hidden-Markov approach of~\cite{robinson2018sequential} further exploit transaction order. Building on this, automated machine learning (AutoML) systems such as Auto-sklearn~\cite{feurer2015efficient}, FLAML~\cite{wang2021flaml}, and AutoGluon~\cite{erickson2020autogluon} search over algorithms and hyperparameters to build gradient-boosted tree ensembles like XGBoost~\cite{chen2016xgboost}, which remains the de facto standard since tree-based models still match or outperform deep networks on tabular data~\cite{grinsztajn2022tree}. However, AutoML explores a fixed numerical search space without reasoning about dataset semantics, performs no domain-grounded feature engineering, and returns a model rather than a human-readable rationale; its search is statistical, not diagnostic. Another paradigm models fraudulent behavior on a graph of users, transactions, and devices~\cite{cheng2024graph}. Representative methods include the semi-supervised graph attention network of~\cite{wang2019semi}, the camouflage-resistant CARE-GNN~\cite{dou2020enhancing}, and the imbalance-aware PC-GNN~\cite{liu2021pick}. These methods excel at group-level and link-level analysis, but require a pre-defined construction graph, entailing costly relational-data collection, substantial entity-resolution difficulty, and manual architecture design while remaining opaque at the individual-decision level. Crucially, in most real-world fraud-screening scenarios the available evidence is plain individual-level tabular data rather than a rich relational graph~\cite{shwartz2022tabular, grover2022fdb}, so the multi-relational structure these methods rely on is often absent or prohibitively expensive to construct. For individual-level classification dominating real deployments, tree-based models tailored to tabular data therefore remain both sufficient and preferable.

\subsection{Gap for Fraud Detection}
\label{sec:rw_llm}

 However, the rise of large language models has intensified the threat of fraud: generative AI enables fraudsters to produce personalized phishing messages and deepfake content at scale, weakening the lexical and grammatical cues traditional detection methods rely on~\cite{schmitt2024digital}. Meanwhile, high-quality classical pipelines require substantial expert effort in feature engineering and tuning, limiting their responsiveness as fraud techniques evolve. More recently, large language models have been used as autonomous agents that reason and act in external environments. AutoGen~\cite{wu2023autogen} orchestrates multiple conversable agents to solve tasks collaboratively; Reflexion~\cite{shinn2023reflexion} reinforces agents through verbal self-reflection rather than weight updates; and textual-gradient methods~\cite{emnlp2023NLG} optimize programs by treating natural-language critiques as gradients. Closer to data work, Data Interpreter~\cite{hong2024data} decomposes data-science tasks into hierarchical plans and invokes tools for modeling, while coding agents such as Claude Code~\cite{anthropic2025claudecode} autonomously write, execute, and debug code from natural-language instructions. These frameworks show that an LLM agent can diagnose, plan, and self-correct in language, but they have been developed on generic reasoning, coding, and data-science benchmarks, with none tailored to fraud detection or to the strict recall and precision constraints of production anti-fraud systems. Table~\ref{tab:related_compare} contrasts the main paradigms against the requirements of individual-level fraud detection. As the table shows, AutoML suits tabular data but is neither semantic nor fraud-specific; graph methods are fraud-oriented but require relational structure and are opaque; and current LLM agents are agentic and interpretable yet not designed for fraud detection. No existing approach simultaneously satisfies all four requirements, which motivates an LLM-driven agent purpose-built for tabular fraud detection.

\begin{table}[!t]
    \centering
    \caption{Representative methods against the four requirements of individual level fraud detection.}
    \label{tab:related_compare}
    \scriptsize
    \setlength{\tabcolsep}{2pt}
    \renewcommand{\arraystretch}{1.15}
    \begin{tabularx}{\columnwidth}{@{}lCccccc@{}}
        \toprule
        \textbf{Method} & \textbf{Paradigm} & \textbf{Agentic} & \textbf{Tabular} & \textbf{Interpretable} & \textbf{Fraud-specific} \\
        \midrule
        Auto-sklearn~\cite{feurer2015efficient} & AutoML    & \textcolor{red!75!black}{\ding{55}} & \textcolor{green!55!black}{\ding{51}} & \textcolor{red!75!black}{\ding{55}} & \textcolor{red!75!black}{\ding{55}} \\
        FLAML~\cite{wang2021flaml}              & AutoML    & \textcolor{red!75!black}{\ding{55}} & \textcolor{green!55!black}{\ding{51}} & \textcolor{red!75!black}{\ding{55}} & \textcolor{red!75!black}{\ding{55}} \\
        AutoGluon~\cite{erickson2020autogluon}  & AutoML    & \textcolor{red!75!black}{\ding{55}} & \textcolor{green!55!black}{\ding{51}} & \textcolor{red!75!black}{\ding{55}} & \textcolor{red!75!black}{\ding{55}} \\
        CARE-GNN~\cite{dou2020enhancing}        & GNN       & \textcolor{red!75!black}{\ding{55}} & \textcolor{red!75!black}{\ding{55}} & \textcolor{red!75!black}{\ding{55}} & \textcolor{green!55!black}{\ding{51}} \\
        PC-GNN~\cite{liu2021pick}               & GNN       & \textcolor{red!75!black}{\ding{55}} & \textcolor{red!75!black}{\ding{55}} & \textcolor{red!75!black}{\ding{55}} & \textcolor{green!55!black}{\ding{51}} \\
        AutoGen~\cite{wu2023autogen}            & LLM agent & \textcolor{green!55!black}{\ding{51}} & \textcolor{green!55!black}{\ding{51}} & \textcolor{green!55!black}{\ding{51}} & \textcolor{red!75!black}{\ding{55}} \\
        Data Interp.~\cite{hong2024data}        & LLM agent & \textcolor{green!55!black}{\ding{51}} & \textcolor{green!55!black}{\ding{51}} & \textcolor{green!55!black}{\ding{51}} & \textcolor{red!75!black}{\ding{55}} \\
        Claude Code~\cite{anthropic2025claudecode} & LLM agent & \textcolor{green!55!black}{\ding{51}} & \textcolor{green!55!black}{\ding{51}} & \textcolor{green!55!black}{\ding{51}} & \textcolor{red!75!black}{\ding{55}} \\
        \bottomrule
    \end{tabularx}
\end{table}

%% file: sections/03_preliminaries.tex
\section{Preliminaries}
\label{sec:preliminaries}

This section establishes the formal foundations on which SAGE is built. We define the tabular fraud detection problem (Section~\ref{sec:prob_formal}), characterize the class imbalance that shapes any fraud detector (Section~\ref{sec:imbalance}), and introduce the modeling and optimization primitives reused throughout Section~\ref{sec:methodology} (Section~\ref{sec:prelim_model}).

\subsection{Problem Formalization}
\label{sec:prob_formal}

We study fraud detection in its tabular, individual-transaction form. Let each instance be a feature vector $\mathbf{x} \in \mathcal{X} \subseteq \mathbb{R}^{p}$ describing a single transaction or account, together with a binary label $y \in \{0, 1\}$, where $y=1$ denotes a fraudulent (positive) instance and $y=0$ a legitimate (negative) one. A dataset $P = (X, y)$ collects $n$ such instances as a feature matrix $X \in \mathbb{R}^{n \times p}$ and a label vector $y \in \{0,1\}^n$. The goal of a fraud detector is to obtain a scoring function
\begin{equation}
    f : \mathcal{X} \to [0, 1], \qquad \hat{y} = \mathbf{1}\!\left[\, f(\mathbf{x}) \geq \delta \,\right],
    \label{eq:scoring}
\end{equation}
that maps each instance to a fraud probability, which a decision threshold $\delta \in [0,1]$ converts into a hard prediction $\hat{y}$. In the remainder of the paper, this scoring function is realized as the model $M \in \mathcal{M}$ produced by SAGE, and the threshold $\delta$ corresponds to the tunable \texttt{threshold\_tune} handle of Section~\ref{sec:planning}. Following \cite{davis2006relationship} and \cite{dalpozzolo2014learned}, we treat fraud detection not as plain binary classification but as a constrained problem: maximizing detection quality subject to a tight false-alarm budget and a minimum-recall requirement, a view that directly motivates the composite reward of Section~\ref{sec:reward}. We use $\delta$ throughout for the decision threshold and reserve $\tau$ (introduced in Section~\ref{sec:prelim_mdp_nlg_reward}) for the dataset-specific recall constraint in the reward function.

\subsection{Class Imbalance}
\label{sec:imbalance}

A defining property of fraud data is extreme class imbalance, which we quantify by the \emph{imbalance ratio}
\begin{equation}
    \mathrm{IR} \;=\; \frac{n_{\mathrm{neg}}}{n_{\mathrm{pos}}},
    \label{eq:ir}
\end{equation}
where $n_{\mathrm{pos}}$ and $n_{\mathrm{neg}}$ are the numbers of positive and negative instances; in real fraud datasets $\mathrm{IR}$ commonly spans two orders of magnitude. Under such skew, a trivial classifier predicting every instance as legitimate attains near-perfect accuracy while detecting no fraud, rendering accuracy uninformative~\cite{lucas2020survey, yousefi2019comprehensive}. Common remedies span \emph{data-level} resampling that rebalances the training distribution (e.g., SMOTE~\cite{chawla2002smote}) and \emph{algorithm-level} loss reweighting such as a positive-class weight $w_{\mathrm{pos}} \approx \mathrm{IR}$.

\subsection{Formal Definitions of SAGE}
\label{sec:prelim_model}

\subsubsection{SAGE as a Composite Mapping}
\label{sec:prelim_sage}

Let $\mathcal{P}$ denote the \emph{problem space}, the set of tabular fraud detection problems $P=(X,y)$ introduced in Section~\ref{sec:prob_formal}. Let $\mathcal{M}$ denote the \emph{model space} of trained fraud classifiers, $\mathcal{C}$ the \emph{code space} of executable training scripts, and $\mathbb{R}^d$ the $d$-dimensional metric vector space (in this work $d=4$, covering AUPRC, F1, MCC, and $\mathrm{R}@\mathrm{FPR}_{10^{-4}}$). SAGE is then formalized as the composite mapping
\begin{align}
    \mathrm{SAGE} &: \mathcal{P} \to \mathcal{C} \times \mathcal{M} \times \mathbb{R}^d, \nonumber \\
    \mathrm{SAGE}(P) &= \mathcal{A}_3\!\big(P,\, \mathcal{A}_2(P,\, \mathcal{A}_1(P))\big),
    \label{eq:sage_overall}
\end{align}
in which the three agents $\mathcal{A}_1, \mathcal{A}_2, \mathcal{A}_3$ are applied strictly in sequence. The final output is the triple $(c^{*}, M^{*}, \mathbf{m}^{*})$ recorded at the best iteration $t^{*} = \arg\max_t R(\mathbf{m}_t)$, where $R(\cdot)$ is the composite reward defined in Eq.~\eqref{eq:reward_main} and $\mathbf{m}_t \in \mathbb{R}^d$ is the metric vector of the model at iteration $t$.

\subsubsection{The structure of Three Agents}
\label{sec:prelim_agents}

Each agent in Eq.~\eqref{eq:sage_overall} is a typed mapping between intermediate spaces:
\begin{align}
    \mathcal{A}_1 &: \mathcal{P} \to \mathcal{R} \times \mathcal{T}, \nonumber \\
    \mathcal{A}_2 &: \mathcal{P} \times \mathcal{R} \times \mathcal{T} \to \mathcal{C} \times \mathcal{M}, \nonumber \\
    \mathcal{A}_3 &: \mathcal{P} \times \mathcal{C} \times \mathcal{M} \to \mathcal{C} \times \mathcal{M} \times \mathbb{R}^d,
    \label{eq:agent_types}
\end{align}
where $\mathcal{R}$ is the profiling report space and $\mathcal{T}$ the DDT space (Section~\ref{sec:prelim_ddt}). Internally, each agent decomposes as follows. The Profiling Agent $\mathcal{A}_1$ runs a deterministic statistical computation $\mathrm{Stats}(\cdot)$, builds the Data Diagnostic Tree, and adds a semantic interpretation through an LLM call:
\begin{equation}
    \begin{split}
        &\mathcal{A}_1(P) = (\rho, T), \\
        &\rho = \mathrm{LLM}_{\mathrm{interpret}}(\mathrm{Stats}(P)), \\
        &T = \mathrm{DDT}(P).
    \end{split}
    \label{eq:agent1_decomp}
\end{equation}
The Planning Agent $\mathcal{A}_2$ then consumes $(\rho, T)$ and synthesizes an initial training program $c_0$, executed in a sandbox runner to obtain the initial model $M_0$ and its metric vector $\mathbf{m}_0$:
\begin{align}
    &\mathcal{A}_2(P, \rho, T) = (c_0, M_0), \nonumber \\
    &c_0 = \mathrm{LLM}_{\mathrm{codegen}}(\rho, T), \nonumber \\
    &M_0 = \mathrm{Execute}(c_0, P).
    \label{eq:agent2_decomp}
\end{align}
Finally, the Optimization Agent $\mathcal{A}_3$ iteratively refines $c_0$ for at most $K$ rounds, returning the triple from the best iteration:
\begin{equation}
    \mathcal{A}_3(P, c_0, M_0) = (c^{*}, M^{*}, \mathbf{m}^{*}), \quad t^{*} = \arg\max_{t \in [0, K]} R(\mathbf{m}_t).
    \label{eq:agent3_decomp}
\end{equation}
Here $\mathrm{LLM}_{\mathrm{interpret}}$ and $\mathrm{LLM}_{\mathrm{codegen}}$ are independent LLM calls, and $\mathrm{Execute}(\cdot, P)$ returns both the trained model and its metric vector $\mathbf{m} \in \mathbb{R}^d$. The structural innovation of $\mathcal{A}_1$ is the DDT $T$ defined next; the optimization loop driving $\mathcal{A}_3$ is formalized in Section~\ref{sec:prelim_mdp_nlg_reward}; the engineering instantiation of each $\mathcal{A}_i$ is presented in Section~\ref{sec:methodology}.

\subsubsection{Data Diagnostic Tree}
\label{sec:prelim_ddt}

The Data Diagnostic Tree $T$ produced by $\mathcal{A}_1$ in Eq.~\eqref{eq:agent1_decomp} is a rooted attributed tree $T = (N, A)$, where $N$ is a set of nodes organized hierarchically with a unique root $n_{\mathrm{root}}$ representing the target dataset, and $A : N \to 2^{\mathcal{K} \times \mathcal{V}}$ assigns to each node a set of key--value pairs drawn from a predefined key space $\mathcal{K}$ and value space $\mathcal{V}$. Every non-root node has exactly one parent. The DDT is structured around six \emph{semantic layers}, each forming one principal branch of the tree:
\begin{equation}
    \begin{split}
        &T = \big\{\, L_{\mathrm{scale}},\; L_{\mathrm{label}},\; L_{\mathrm{feature}},\\
        &\qquad\; L_{\mathrm{quality}},\; L_{\mathrm{structure}},\; L_{\mathrm{diagnosis}}\,\big\}.
    \end{split}
    \label{eq:ddt_layers}
\end{equation}
The encoded content of each layer and the procedure that constructs $T$ are detailed in Section~\ref{sec:profiling}.

\subsubsection{MDP, Natural-Language Gradient, and Reward}
\label{sec:prelim_mdp_nlg_reward}

The iterative code refinement of $\mathcal{A}_3$ in Eq.~\eqref{eq:agent3_decomp} is cast as a finite-horizon Markov Decision Process~\cite{DBLP2018MDP}, a tuple $\langle \mathcal{S}, \mathcal{A}, \mathcal{T}_{\mathrm{trans}}, R, K \rangle$ with state space $\mathcal{S}$, action space $\mathcal{A}$, transition $\mathcal{T}_{\mathrm{trans}}: \mathcal{S} \times \mathcal{A} \to \mathcal{S}$, reward $R: \mathcal{S} \to \mathbb{R}$, and horizon $K$. The state $s_t$ at iteration $t$ is
\begin{equation}
    s_t = (c_t,\, \mathbf{m}_t,\, H_t),
    \label{eq:state}
\end{equation}
where $c_t \in \mathcal{C}$ is the current code, $\mathbf{m}_t = \mathrm{Eval}(M_t) \in \mathbb{R}^d$ is the metric vector of $M_t = \mathrm{Execute}(c_t, P)$ evaluated on the held-out validation split (Section~\ref{sec:metrics}), and $H_t = \{(a_i, R_i, D_i)\}_{i=1}^{t-1}$ records past actions, rewards, and natural-language diagnoses (with $H_1 = \emptyset$). The action space $\mathcal{A}$ consists of seven handle-targeted edits: \texttt{missing\_strategy}, \texttt{feature\_transform}, \texttt{regularization}, \texttt{depth\_and\_estimators}, \texttt{sampling\_params}, \texttt{threshold\_tune}, and \texttt{algo\_switch}.

Following ReAct~\cite{iclr2023React} and textual-gradient methods~\cite{emnlp2023NLG}, each iteration decomposes into two independent, stateless LLM calls:
\begin{equation}
    \begin{split}
        &D_t = \mathrm{LLM}_{\mathrm{critique}}(c_t, \mathbf{m}_t, H_t), \\
        &c_{t+1} = \mathrm{LLM}_{\mathrm{codegen}}(c_t, D_t, a_t).
    \end{split}
    \label{eq:nlg_pair}
\end{equation}
The first call (\emph{reasoning}) emits a natural-language diagnosis $D_t \in \mathcal{D}$; the second call (\emph{acting}) translates $D_t$, together with the selected action $a_t \in \mathcal{A}$, into a concrete code edit. $D_t$ is the natural-language analog of a numerical gradient: while the latter specifies the descent direction in continuous parameter space, $D_t$ specifies the edit direction in discrete code space.

Finally, the scalar reward $R(\mathbf{m}_t)$ that drives the selection of $t^{*}$ in Eq.~\eqref{eq:agent3_decomp} consolidates the multi-objective requirements of fraud detection:
\begin{equation}
\resizebox{\linewidth}{!}{$
    R(\mathbf{m}_t) = \underbrace{\big(w_1 \mathrm{F1} + w_2 \mathrm{AUPRC} + w_3 \mathrm{R}@\mathrm{FPR}_{10^{-4}} + w_4 \mathrm{Recall}\big)}_{\text{primary score}} \cdot W_{\mathrm{recall}} \cdot W_{\mathrm{precision}} \,+\, B,
$}
\label{eq:reward_main}
\end{equation}
where $(w_1, w_2, w_3, w_4) = (0.40, 0.30, 0.20, 0.10)$. The two multiplicative gates and additive bonus realize three fraud-specific design intents: $W_{\mathrm{recall}}$ penalizes recall below a dataset-specific threshold $\tau$ produced by $\mathcal{A}_1$; $W_{\mathrm{precision}}$ prevents degeneration into a trivial high-recall classifier; and $B$ adds a small bonus when F1 or recall exceeds the initial baseline $\mathbf{m}_0$ produced by $\mathcal{A}_2$, providing a non-zero gradient even when both gates saturate. The operational consequences of each term are discussed in Section~\ref{sec:reward}.

%% file: sections/04_methodology.tex
\section{SAGE}
\label{sec:methodology}

\begin{figure*}[!t]
    \centering
    \includegraphics[width=0.95\textwidth]{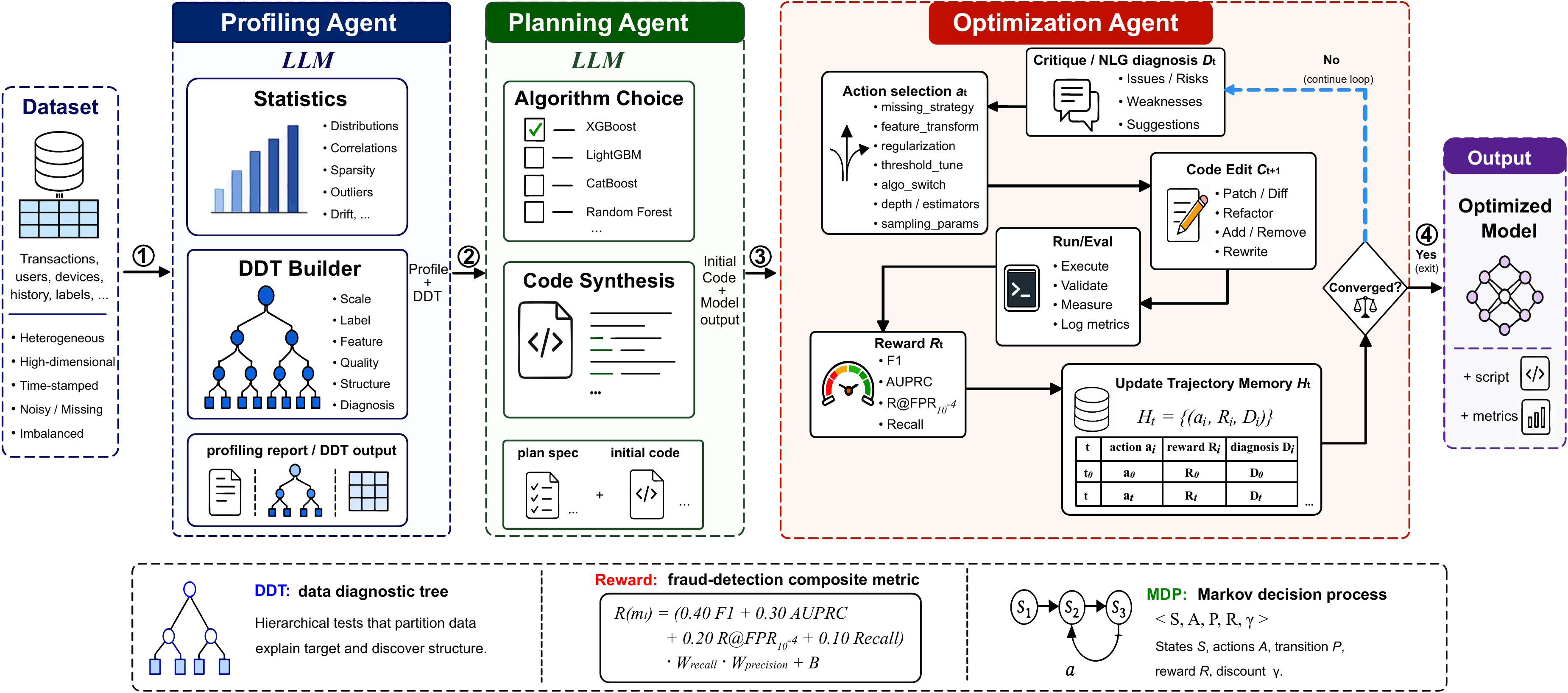}
    \caption{Overall architecture of the SAGE framework. \ding{172} SAGE takes a heterogeneous tabular fraud dataset as input, \ding{173} profiles it through statistical analysis and a six-layer Data Diagnostic Tree (DDT), \ding{174} and uses the resulting data diagnosis to select an algorithm and synthesize the initial training code. \ding{175} The Optimization Agent then iteratively refines the code through an NLG-guided MDP loop, where code edits are evaluated by a fraud-specific composite reward and stored in trajectory memory. The process stops after convergence and returns the optimized model, final script, and evaluation metrics.}
    \label{fig:sage_architecture}
\end{figure*}


SAGE balances LLM \emph{autonomy} (each agent independently chooses algorithms, designs features, diagnoses bottlenecks, and edits code) with explicit \emph{mathematical constraints} (the DDT grounds early decisions in data, the finite-horizon MDP~\cite{DBLP2018MDP} with natural-language gradient~\cite{emnlp2023NLG} signals constrains the iterative loop, and a composite reward enforces fraud-specific objectives), enabling SAGE to run without human intervention. SAGE decomposes the end-to-end process into three specialized, collaborative LLM-driven agents (Eq.~\eqref{eq:sage_overall}), each responsible for a classic operation in the fraud detection workflow. The \textbf{Profiling Agent} ($\mathcal{A}_1$, Section~\ref{sec:profiling}) examines the input dataset and generates a data-aware profiling result containing statistical reports and a DDT. The \textbf{Planning Agent} ($\mathcal{A}_2$, Section~\ref{sec:planning}) uses this profiling result to select an appropriate algorithm and synthesize an initial executable training program tailored to the dataset. The \textbf{Optimization Agent} ($\mathcal{A}_3$, Section~\ref{sec:optimization}) then iteratively optimizes the program in the code space through an NLG-guided MDP under a fraud-specific reward, until convergence. These three agents constitute a strictly sequential process (Figure~\ref{fig:sage_architecture}). Communication flows forward, with each downstream agent relying on the structured output of the upstream agent. The user supplies only the dataset $P$ and the target label column; the entire pipeline proceeds without human intervention and returns the triple $(c^{*}, M^{*}, \mathbf{m}^{*})$ at the best iteration.

\subsection{Profiling Agent with DDT}
\label{sec:profiling}

The Profiling Agent $\mathcal{A}_1$ is the entry point of SAGE. Following the typed signature in Eq.~\eqref{eq:agent1_decomp}, it transforms a raw fraud detection problem $P=(X,y)$ into a structured, fraud-aware diagnosis $(\rho, T)$ that grounds every subsequent decision in the data itself. We focus here on the engineering realization of the DDT, which is the central structural innovation of $\mathcal{A}_1$.

\subsubsection{Why a tree}
\label{sec:ddt_def}

Feeding LLMs with raw fraud datasets exposes three coupled bottlenecks: (i) \emph{information overload}, where hundreds of raw feature descriptors dilute key signals; (ii) \emph{semantic gap}, where raw statistics (e.g., \texttt{imbalance\_ratio}$=577.3$) require an additional reasoning step to become actionable; and (iii) \emph{narrow attention}, where LLMs overemphasize the most salient features while neglecting global structure. The Data Diagnostic Tree (DDT), formally defined in Eq.~\eqref{eq:ddt_layers}, addresses all three by compressing the dataset into a compact, semantically annotated tree of six layers. Table~\ref{tab:ddt_layers} summarizes the encoded content and downstream usage of each layer.

\begin{table}[!t]
    \centering
    \caption{The six semantic layers of the Data Diagnostic Tree (DDT).}
    \label{tab:ddt_layers}
    \scriptsize
    \setlength{\tabcolsep}{3pt}
    \begin{tabularx}{\columnwidth}{@{}l X@{}}
        \toprule
        \textbf{Layer} & \textbf{Encoded content} \\
        \midrule
        $L_{\mathrm{scale}}$ & Rows, features, memory footprint \\
        $L_{\mathrm{label}}$ & Fraud rate, imbalance ratio, positive/negative counts \\
        $L_{\mathrm{feature}}$ & Numeric/categorical counts, top label-correlated features \\
        $L_{\mathrm{quality}}$ & Missing columns, high-missing columns, redundant pairs \\
        $L_{\mathrm{structure}}$ & ID-like columns, time columns, subset-fraud columns \\
        $L_{\mathrm{diagnosis}}$ & Difficulty, dataset type, key risk features, recall threshold $\tau$ \\
        \bottomrule
    \end{tabularx}
\end{table}

Downstream agents receive the DDT as indented hierarchical text, in which each of the six branches contains raw statistical fields generated by $\mathrm{Stats}(P)$ together with semantic annotations, with the origin of each annotation (\emph{rule-based} or \emph{LLM-derived}) marked inline. This inline origin tag is critical: it allows downstream agents to weigh deterministic facts and LLM inferences differently when their conclusions disagree, and lets human auditors trace every diagnostic claim back to either a numerical rule or a specific LLM call. Separating these two origins also enables targeted debugging: an incorrect statistic reflects a flaw in $\mathrm{Stats}(P)$, whereas an unreasonable diagnosis points to the $\mathrm{LLM}_{\mathrm{interpret}}$ prompt or backbone. The indented format itself is deliberately chosen to align with the token-level patterns LLMs have absorbed from code and configuration files during pre-training, allowing the agent to traverse the six branches with minimal parsing overhead and avoiding the verbosity of JSON or XML serialization that would otherwise inflate the prompt budget. The same structure remains stable across datasets of vastly different scale, so prompt templates downstream of $\mathcal{A}_1$ do not need to be rewritten for each new fraud domain. This representation compresses the statistical fingerprint of IEEE-CIS, a flattened report of over $50{,}000$ tokens covering all $433$ features, into a six-layer tree of approximately $2{,}000$ tokens, a $25\times$ reduction while preserving the decision-critical signals required by downstream agents. The same compression trend holds on the other four datasets, with reductions ranging from $4\times$ on the lower-dimensional Credit Card data to over $20\times$ on the high-dimensional industrial telecom dataset. A partial excerpt for the IEEE-CIS dataset is shown below; note that values such as $\tau=0.75$ are produced by $\mathrm{LLM}_{\mathrm{interpret}}$ from dataset-specific semantics rather than hard-coded defaults:
\begin{lstlisting}[basicstyle=\ttfamily\scriptsize, frame=single, framesep=3pt, xleftmargin=4pt]
DDT("ieee_cis")
|- Scale
|   |- n_rows: 590,540
|   |- n_features: 433
|   +- note: large-scale dataset
|- Label
|   |- imbalance_ratio: 27.58
|   +- difficulty [rule]: HARD
|- Feature
|   |- n_numeric: 400, n_cat: 33
|   +- top_correlated
|       |- V258 (corr = 0.42)
|       +- V201 (corr = 0.38)
|- Quality
|   +- note: heavy missingness,
|      native-NA preferred
|- Structure
|   +- note: identifier-only,
|      no temporal partition
+- Diagnosis [LLM]
    |- type: transactional fraud
    |- key_risk: [V258, V201, C14]
    +- recall_threshold: tau=0.75
\end{lstlisting}

\subsubsection{DDT Construction Procedure}
\label{sec:ddt_proc}

The DDT is constructed in two phases, separating statistical numerical analysis from LLM-based semantic interpretation.

\paragraph{Phase~1: Statistical Computation $\mathrm{Stats}(P)$ ($L_{\mathrm{scale}}$--$L_{\mathrm{structure}}$)}
This phase is entirely rule-based. Given $P$, $\mathrm{Stats}(P)$ computes the basic dimensions, the imbalance ratio $\mathrm{IR}=n_{\mathrm{neg}}/n_{\mathrm{pos}}$, feature types, missing rate per column ($50\%$ high-missingness threshold, following common practice in tabular preprocessing), distribution moments (skewness, kurtosis), top-$K$ feature--label correlations, and special structure detection (ID columns with uniqueness $>95\%$, time columns, subset-fraud columns). Each statistic is mapped to semantic labels through threshold rules and written into the corresponding branch (e.g., $\mathrm{IR}\!>\!100$ becomes ``extremely imbalanced, requiring sample weighting'' under $L_{\mathrm{label}}$), populating the first five layers.

\paragraph{Phase~2: Diagnostic Inference ($L_{\mathrm{diagnosis}}$)}
The sixth layer is populated by a single $\mathrm{LLM}_{\mathrm{interpret}}$ call that reasons over the first five layers and outputs: \textbf{(i) Dataset type} (transactional, behavioral, or profile-based fraud), inferred from feature names, top-correlated features, and the presence of time/ID columns; \textbf{(ii) Key risk features}, a short list of fraud-distinguishing columns drawn from the top-$K$ correlated columns combined with semantic priors of feature names (e.g., transaction amount, device fingerprint, time field); and \textbf{(iii) Recall threshold $\tau$}, subsequently used by the recall-constraint gate $W_{\mathrm{recall}}$ in Eq.~\eqref{eq:reward_main}. By committing to a complete dataset profile at this analysis phase, rather than letting downstream agents drift on arbitrary cues, SAGE remains grounded in dataset evidence throughout the optimization loop.

\subsection{Planning Agent}
\label{sec:planning}

The Planning Agent $\mathcal{A}_2$ converts the structured diagnosis $(\rho, T)$ produced by $\mathcal{A}_1$ into an executable training program tailored to the target dataset, following Eq.~\eqref{eq:agent2_decomp}. $\mathcal{A}_2$ delegates almost the entire generation task to a single $\mathrm{LLM}_{\mathrm{codegen}}$ call, with algorithm selection, feature engineering, hyperparameter values, and thresholding logic all autonomously written by the LLM. This autonomy is safe rather than chaotic, because the structured DDT $T$ ensures it is grounded in dataset evidence.

\subsubsection{Algorithm Selection Logic}
\label{sec:algo_selection}

The core responsibility of $\mathcal{A}_2$ is to select an algorithm family for $c_0$. Here, we deliberately avoid using hard-coded selection rules. Instead, leveraging the inference capabilities of the LLM-driven agent, $\mathrm{LLM}_{\mathrm{codegen}}$ receives the complete DDT $T$ in its prompt. Our prompt instructs the LLM to weigh the $L_{\mathrm{scale}}$, $L_{\mathrm{label}}$, $L_{\mathrm{feature}}$, and $L_{\mathrm{quality}}$ layers of $T$, evaluate a set of common candidate models against these layers, and describe the empirical strengths of each candidate. After committing to a choice, the LLM is also required to articulate why the selected algorithm is best suited for the diagnosed scenario. An example of the SAGE built-in prompt used at this stage is shown below:

\begin{tcolorbox}[
    enhanced,
    colback=white,
    colframe=black,
    boxrule=0.6pt,
    arc=2mm,
    left=8pt, right=8pt, top=8pt, bottom=8pt,
    fontupper=\small,
    overlay={
        \node[
            fill=black, text=white, font=\small\bfseries\itshape,
            inner xsep=6pt, inner ysep=2pt, rounded corners=2pt,
            anchor=west
        ] at ([xshift=10pt, yshift=0pt]frame.north west) {Prompt for $\mathrm{LLM}_{\mathrm{codegen}}$};
    },
    before skip=8pt, after skip=8pt
]
\vspace{4pt}
You are an expert ML engineer for fraud detection. You will receive a Data Diagnostic Tree (DDT) describing the target dataset and must (i) choose a suitable algorithm, (ii) generate a complete Python training code following the structure (Loading, Preprocessing, Feature Engineering, Training, Metric Computation), and (iii) provide a brief justification for your algorithm choice.

\medskip
\textbf{DDT (input):}\\
\texttt{<DDT omitted for brevity>}

\medskip
\textbf{Empirical cues:}
\begin{itemize}\setlength\itemsep{0pt}
    \item \textbf{XGBoost} --- robust missing-value handling; strong on heterogeneous tabular features.
    \item \textbf{LightGBM} --- high training efficiency on large-scale datasets ($>$1M rows).
    \item \textbf{CatBoost} --- native categorical encoding; strong with high-cardinality categoricals.
    \item $\cdots$ (additional candidates may be appended via prompt extension)
\end{itemize}

\medskip
\textbf{Selection guidelines:} Read $L_{\mathrm{scale}}$ for size, $L_{\mathrm{label}}$ for imbalance, $L_{\mathrm{feature}}$ for feature typing, and $L_{\mathrm{quality}}$ for missingness. Weigh these jointly; a single layer should not dominate.

\medskip
\textbf{Output format:} JSON with fields \texttt{algorithm}, \texttt{code}, \texttt{justification}.
\end{tcolorbox}

\subsubsection{Code Structure Design}
\label{sec:code_structure}

Code $c_0$ must have a fixed five-block structure: data loading, preprocessing, feature engineering, model training, and metric calculation. The specific content of each block is filled in by the LLM, but the block boundaries themselves are immutable. This design provides a consistent working environment for the LLM, reduces quality variation in the generated code, and allows the Agent to intervene and optimize without rewriting the entire script; these locations are referred to as \emph{optimization handles}. Each annotation marked \texttt{\# handle: <name>} corresponds to an action type in the action space $\mathcal{A}$ (defined in Section~\ref{sec:optimization}). These handle conventions enable the natural-language gradients of the Optimization Agent to translate textual information such as ``the threshold for recalling the target is too high'' into precise local edits at the corresponding handle. 

A typical $c_0$ generated by $\mathcal{A}_2$ takes the following form:

\begin{algorithm}[t]
\caption{Initial Training Script $c_0$ by $\mathcal{A}_2$}
\label{alg:c0_skeleton}
\begin{algorithmic}[1]
\Require $P=(X, y)$, $\rho$, DDT $T$
\Ensure $M_0$, $\mathbf{m}_0$
\vspace{2pt}
\State \block{B1: Loading}
\State $df \gets \textsc{Load}(P)$
\vspace{2pt}
\State \block{B2: Preprocessing}
\State $df \gets \textsc{Impute}(df)$
\vspace{2pt}
\State \block{B3: Feature Eng.}
\State $df \gets \textsc{Transform}(df, T)$
\vspace{2pt}
\State \block{B4: Training}
\State $\theta \gets \{\texttt{n\_est},\texttt{depth},\texttt{reg},\ldots\}$
\State $M_0 \gets \textsc{Fit}(\theta, X_{\text{tr}}, y_{\text{tr}})$ 
\vspace{2pt}
\State \block{B5: Evaluation}
\State $\delta \gets 0.5$
\State $\mathbf{m}_0 \gets \textsc{Eval}(M_0, X_{\text{val}}, y_{\text{val}}, \delta)$
\State \Return $(M_0, \mathbf{m}_0)$
\end{algorithmic}
\end{algorithm}

\subsection{NLG-Guided MDP Optimization Agent}
\label{sec:optimization}

The Optimization Agent $\mathcal{A}_3$ iteratively refines the initial code $c_0$ produced by $\mathcal{A}_2$, following Eq.~\eqref{eq:agent3_decomp}, until convergence. The dependence of $\mathcal{A}_3$ on $M_0$ provides the initial baseline metrics $\mathbf{m}_0$ used by the baseline-bonus term $B$ of the reward function (Eq.~\eqref{eq:reward_main}). The loop is implemented as a ReAct-style reasoning--acting cycle~\cite{iclr2023React}: in each iteration, the LLM first performs inference through NLG text diagnostics, then converts the diagnosis into a code edit.

\subsubsection{MDP Loop and Convergence}
\label{sec:mdp_formulation}

The optimization is a finite-horizon MDP whose state $s_t = (c_t, \mathbf{m}_t, H_t)$ and seven handle-targeted actions $\mathcal{A}$ are defined in Eq.~\eqref{eq:state} and Section~\ref{sec:prelim_mdp_nlg_reward}. We focus here on the engineering of the loop. Each tuple $(a_i, R_i, D_i)$ recorded in $H_t$ captures the action selected at iteration $i$, the natural-language diagnosis $D_i$ that motivated it, and the scalar reward $R_i = R(\mathbf{m}_{i+1})$ observed after executing the resulting code $c_{i+1}$. Although the LLM calls within an iteration are stateless, $H_t$ is fed into the next iteration's prompt, granting the agent access to its own trajectory. Action selection defaults to LLM-driven decision-making, but when F1 stagnates for $\geq 2$ consecutive rounds an action switch is triggered, guiding the LLM to explore new actions in $\mathcal{A}$. The loop terminates either when $t = K$, or when $R(\mathbf{m}_t)$ improves by less than a tolerance $\varepsilon$ for $k$ consecutive iterations. We use $K=20$, $k=4$, and $\varepsilon=0.001$ in this work.

\subsubsection{NLG as Textual Gradient}
\label{sec:nlg}

A distinguishing feature of $\mathcal{A}_3$ is that its optimization does not rely solely on metric values to drive code refinement. Each iteration decomposes into two stateless LLM calls (Eq.~\eqref{eq:nlg_pair}): the first call acts as a \emph{critic} that produces a natural-language gradient $D_t$ in the code space, identifying bottlenecks and proposing improvement directions; the second call acts as an \emph{executor} that translates $D_t$, together with the selected action $a_t$, into a concrete code edit at the corresponding handle. The two calls share no memory; only $D_t$ and $a_t$ flow between them, making every reasoning--acting cycle auditable through $H_t$. SAGE's \emph{self-reflection} thus materializes through $H_t$: the LLM reviews its own previous code and outcomes before deciding the next action.

\subsubsection{Composite Reward Design}
\label{sec:reward}

The composite reward $R(\mathbf{m}_t)$ defined in Eq.~\eqref{eq:reward_main} consolidates the multi-objective requirements of fraud detection into a single scalar that drives the MDP. The weights $(w_1, w_2, w_3, w_4) = (0.40, 0.30, 0.20, 0.10)$ reflect the relative priorities of classification balance, ranking quality, low-false-positive-rate detection, and overall recall, while the two multiplicative gates and the additive bonus realize three fraud-specific design intents: $W_{\mathrm{recall}}$ penalizes recall below the dataset-specific threshold $\tau$ set by $\mathcal{A}_1$, $W_{\mathrm{precision}}$ prevents degeneration into a trivially high-recall classifier, and $B$ provides a non-zero gradient signal when both gates saturate. Recall and Precision are computed internally for the reward and are not part of the reported metric vector $\mathbf{m}$; conversely, MCC belongs to $\mathbf{m}$ for final reporting but is excluded from the reward, since F1 and MCC are strongly correlated under extreme imbalance and including both would be redundant for driving optimization. Together, $R$, $D_t$, and $\mathcal{A}$ form a closed optimization loop: $\mathcal{A}_3$ measures progress through $R$, articulates the cause of any shortfall through $D_t$, and acts on a specific handle through $a_t$. Crucially, every quantity inside $R$, F1, AUPRC, R@FPR$_{10^{-4}}$, Recall, Precision, the baseline-bonus comparison, and the threshold value used by \texttt{threshold\_tune}, is computed on the validation split alone, so the test split remains untouched throughout the optimization loop.

\subsection{Algorithm and Complexity}
\label{sec:algorithm}

We now consolidate the three agents into a single end-to-end procedure (Algorithm~\ref{alg:sage_full}) and analyze its computational complexity.

\begin{algorithm}[t]
\caption{SAGE: End-to-End Pipeline}
\label{alg:sage_full}
\begin{algorithmic}[1]
\Require $P$; $K$; $\varepsilon$; $k$
\Ensure $(c^{*}, M^{*}, \mathbf{m}^{*})$
\vspace{2pt}
\State \block{Stage 1:} $\mathcal{A}_1$ Profiling
\State $\rho \gets \textsc{Interpret}(\textsc{Stats}(P))$
\State $T \gets \textsc{BuildDDT}(P)$ \Comment{6-layer}
\vspace{2pt}
\State \block{Stage 2:} $\mathcal{A}_2$ Planning
\State $c_0 \gets \textsc{Codegen}(\rho, T)$
\State $(M_0, \mathbf{m}_0) \gets \textsc{Execute}(c_0, P)$
\vspace{2pt}
\State \block{Stage 3:} $\mathcal{A}_3$ Optimization
\State $H_1 \gets \emptyset$; $t \gets 1$; $(c_t, M_t, \mathbf{m}_t) \gets (c_0, M_0, \mathbf{m}_0)$
\While{$t < K$ \textbf{and} not converged}
    \State $D_t \gets \textsc{Critique}(c_t, \mathbf{m}_t, H_t)$ \Comment{reasoning}
    \State $a_t \gets \pi(s_t)$ \Comment{LLM-driven; force-explore on stagnation}
    \State $c_{t+1} \gets \textsc{Codegen}(c_t, D_t, a_t)$ \Comment{acting}
    \State $(M_{t+1}, \mathbf{m}_{t+1}) \gets \textsc{Execute}(c_{t+1}, P)$
    \State $R_t \gets R(\mathbf{m}_{t+1})$ \Comment{Eq.~\eqref{eq:reward_main}}
    \State $H_{t+1} \gets H_t \cup \{(a_t, R_t, D_t)\}$; $t \gets t{+}1$
\EndWhile
\State $t^{*} \gets \arg\max_{t} R(\mathbf{m}_t)$
\State $(c^{*}, M^{*}, \mathbf{m}^{*}) \gets (c_{t^{*}}, M_{t^{*}}, \mathbf{m}_{t^{*}})$
\State \Return $(c^{*}, M^{*}, \mathbf{m}^{*})$
\end{algorithmic}
\end{algorithm}

\paragraph{Complexity analysis}
We characterize the cost of each stage in terms of two atomic operations: one LLM call $\mathcal{O}(L)$, whose latency depends primarily on the model's response time to the prompt, and one sandboxed code execution $\mathcal{O}(E)$, which trains the model on $P$.

$\mathcal{A}_1$ costs $\mathcal{O}(np + L)$ from a statistical traversal of $P$ plus one LLM call; $\mathcal{A}_2$ costs $\mathcal{O}(L+E)$ from one LLM call and one sandbox execution; and $\mathcal{A}_3$ costs $\mathcal{O}(K(L+E))$ from at most $K$ iterations each consuming two LLM calls and one execution. The total cost of SAGE is therefore
\begin{equation}
    \mathcal{O}\!\big(\underbrace{np}_{\text{Stats}} + \underbrace{(K+1)\, L}_{\text{LLM calls}} + \underbrace{(K+1)\, E}_{\text{executions}}\big) \;=\; \mathcal{O}\!\big(K\,(L + E)\big),
    \label{eq:complexity}
\end{equation}
given $K \gg 1$ and $np \ll K(L+E)$. Empirically, each LLM call ($L$) takes 5--30 seconds on commercial APIs, while each sandboxed model training ($E$) takes seconds to minutes depending on dataset size. In our full-pipeline runs, the average end-to-end wall-clock time is dominated by the Stage~3 Optimization Agent, consistent with the linear-in-$K$ scaling predicted by Eq.~\eqref{eq:complexity}.

%% file: sections/05_experiments.tex
\section{Experiments}
\label{sec:experiments}

\subsection{Experimental Setup}
\label{sec:exp_setup}

\subsubsection{Datasets}
We evaluated SAGE using five tabular fraud detection datasets. Four of these datasets are publicly available and widely used fraud benchmark datasets from the Kaggle platform; the fifth dataset, named TeleGuard, is a non-public dataset provided by a telecommunications operator we are collaborating with. Table~\ref{tab:datasets} summarizes their key statistics.


\begin{table}[!t]
    \centering
    \caption{Statistics of the five fraud detection datasets. IR = imbalance ratio ($n_{\mathrm{neg}}/n_{\mathrm{pos}}$).}
    \label{tab:datasets}
    \scriptsize
    \setlength{\tabcolsep}{3pt}
    \resizebox{\linewidth}{!}{
    \begin{tabular}{llllll}
        \toprule
        \textbf{Dataset} & \textbf{\#Rows} & \textbf{\#Feat.} & \textbf{Fraud} & \textbf{IR} & \textbf{Domain} \\
        \midrule
        Credit Card & 284{,}807     & 30  & 0.17\% & 577.9 & Credit-card \\
        PaySim      & 6{,}362{,}620 & 10  & 0.13\% & 773.7 & Mobile-payment \\
        IEEE-CIS    & 590{,}540     & 433 & 3.50\% & 27.58 & E-commerce \\
        Elliptic    & 46{,}564      & 166 & 9.76\% & 9.2   & Bitcoin \\
        TeleGuard   & 31{,}130      & 47  & 1.00\% & 99.1  & Telecom \\
        \bottomrule
    \end{tabular}
    }
\end{table}

The five datasets together span a wide spectrum of fraud detection scenarios. \textbf{Credit Card}\footnote{\url{https://www.kaggle.com/datasets/mlg-ulb/creditcardfraud}} contains 28 PCA-anonymized principal components plus the raw \texttt{Amount} and \texttt{Time} columns, collected over two days of European credit-card transactions in September 2013~\cite{dal2015calibrating}. \textbf{PaySim}\footnote{\url{https://www.kaggle.com/datasets/ealaxi/paysim1}} is a synthetic mobile-payment log simulated from real African operator data, covering five transaction types (\texttt{CASH\_IN}, \texttt{CASH\_OUT}, \texttt{TRANSFER}, \texttt{DEBIT}, \texttt{PAYMENT})~\cite{lopezrojas2016paysim}. \textbf{IEEE-CIS}\footnote{\url{https://www.kaggle.com/c/ieee-fraud-detection}} is the e-commerce fraud benchmark released by Vesta Corporation, containing 433 anonymized features (the V1--V339 block) with missing rates exceeding $75\%$ in many columns. \textbf{Elliptic}\footnote{\url{https://www.kaggle.com/datasets/ellipticco}} is a Bitcoin transaction table annotated with licit/illicit labels; each column carries 166 features combining transaction attributes and graph-aggregated statistics~\cite{weber2019antimoney}. At last, \textbf{TeleGuard} is a non-public dataset provided by our industrial collaborator, a telecom operator. It comprises 47 features covering subscriber profiles, calling behavior, account-opening patterns, and network metadata, with each record labeled by the operator's anti-fraud investigation team. TeleGuard is included not to expand the public benchmark suite, but to verify that SAGE remains effective on a real industrial dataset whose distribution and feature semantics differ substantially from those of public benchmarks.

\subsubsection{Baselines}
\label{sec:baselines}

We compared SAGE to the following five baselines:

\begin{itemize}\setlength\itemsep{2pt}
    \item \textbf{RS (Random Search)} --- For each dataset, RS adopts the same algorithm, data preprocessing, and feature engineering that SAGE selects, and then performs random sampling over the hyperparameter space of that algorithm. The number of sampling rounds is set equal to the number of optimization iterations SAGE uses on the same task, and the best-performing round is recorded as the result. RS thus replaces SAGE's directed, gradient-like refinement with undirected random search under an identical search budget.
    \item \textbf{Manual (3 human data scientists)} --- We invited three anti-fraud data scientists to manually perform data analysis, feature engineering, model selection, and hyperparameter tuning for each dataset, thereby establishing a robust human benchmark and achieving average-level results. This reflects the workload and output of a typical industry anti-fraud expert team without the use of automation.
     \item \textbf{FLAML}~\cite{wang2021flaml} --- Microsoft's open-source AutoML framework that uses cost-aware Bayesian optimization to automatically select algorithms and tune hyperparameters within a given time budget. Rather than treating all configurations as equally expensive, FLAML explicitly models the training cost of each candidate and prioritizes low-cost, high-potential configurations, making it a strong and efficiency-oriented representative of classical AutoML. We grant FLAML a time budget comparable to SAGE's end-to-end runtime to ensure a fair comparison.
    \item \textbf{AutoGluon}~\cite{erickson2020autogluon} --- Amazon's open-source AutoML framework that builds an ensemble of multiple models via bagging and multi-layer stacking, returning the strongest combined predictor. Unlike methods that search for a single best model, AutoGluon trains a diverse pool of base learners (including gradient-boosted trees, neural networks, and others) and combines them through a layered stacking strategy, which typically yields highly competitive accuracy at the cost of greater training and inference overhead. It represents the ensemble-based, accuracy-oriented state of the art in AutoML, and serves as the strongest non-LLM automated baseline in our comparison.
    \item \textbf{LLM as Coder} --- a single-call LLM baseline. We pass the entire dataset description directly to the LLM and ask it to generate the complete model code and execute it in one shot, obtaining the result. We use Claude Opus~4.7~\cite{anthropic2026claude}, a powerful LLM
    as both the data analyst and the coder. This baseline uses no multi-agent design, no DDT structure, and no iterative optimization; it represents the simplest form of LLM-driven coding.
\end{itemize}

All baselines operate under the same three-way data split as SAGE (Section~\ref{sec:metrics}). RS selects its best-performing round, FLAML and AutoGluon tune internally, Manual experts choose their final hyperparameters and threshold, and LLM-as-Coder runs its single-shot evaluation --- in every case, model selection and threshold tuning are performed on the validation split, and the reported test-split metrics are not used at any selection stage.

\subsubsection{Evaluation Metrics}
\label{sec:metrics}

Fraud detection is a severely imbalanced classification task with multi-objective operational constraints, so single-number metrics such as accuracy are uninformative. We will use the following four indicators as our main reporting indicators:

\begin{itemize}\setlength\itemsep{2pt}
    \item \textbf{F1} --- the harmonic mean of precision and recall, measuring the overall balance between catching more fraud and reducing false alarms.
    \item \textbf{MCC (Matthews Correlation Coefficient)}~\cite{matthews1975comparison} --- a correlation-based classification metric that takes values in $[-1, 1]$ and remains informative under extreme class imbalance; $\text{MCC}=1$ denotes perfect classification and $\text{MCC}=0$ denotes random prediction.
    \item \textbf{R@FPR$_{10^{-4}}$ (Recall at FPR$=10^{-4}$)} --- the recall achieved when the false-positive rate is fixed at one in ten thousand, capturing the model's detection ability under the stringent false-alarm budget typical of production fraud systems. On datasets whose test split contains fewer than $10{,}000$ negative samples, this operating point falls below the minimum measurable FPR.
    \item \textbf{AUPRC (Area Under the Precision--Recall Curve)}~\cite{davis2006relationship} --- the threshold independent ranking quality, widely regarded as the most informative scalar metric for imbalanced binary classification.
\end{itemize}

\paragraph{Data splits}
For every dataset we use a strict three-way split: the data is first partitioned into a training set (70\%), a validation set (10\%), and a test set (20\%) under each random seed. The validation set is used exclusively inside SAGE's optimization loop: the metric vector $\mathbf{m}_t$ that drives the reward $R(\mathbf{m}_t)$ and the best-iteration selection $t^{*} = \arg\max_t R(\mathbf{m}_t)$ are computed on the validation set, never on the test set. The action \texttt{threshold\_tune} likewise sweeps the decision threshold $\delta$ on the validation set only. All reported metrics in Section~\ref{sec:main_results} (F1, MCC, R@FPR$_{10^{-4}}$, AUPRC) are computed on the held-out test set using the model $M^{*}$ and threshold $\delta^{*}$ selected on the validation set, ensuring that no test data participates in any model selection or hyperparameter tuning. The same three-way split is applied identically to all baselines (RS, Manual, FLAML, AutoGluon, LLM-as-Coder); each baseline tunes on the validation set and reports on the test set under the matching seed.

\paragraph{Seed protocol}
Each dataset is evaluated under five random seeds (42, 123, 456, 789, 2024). The seed simultaneously controls four sources of randomness: (i) the train/val/test partition, (ii) model initialization, (iii) sampling-based imbalance handlers (SMOTE, class weights), and (iv) LLM token sampling via the API-level \texttt{seed} parameter. All reported results are the mean and standard deviation over these five seeds. To prevent target leakage, all data-dependent preprocessing statistics, including frequency encodings, target encodings, mean imputation values, and feature standardization parameters, are fit on the training split alone and applied as a frozen transformation to the validation and test splits.

\subsubsection{Implementation Details}
\label{sec:impl}

\paragraph{LLM backbones}
We instantiate SAGE with five different LLM backbones to demonstrate that the framework is not tied to any particular vendor: Claude Opus~4.7~\cite{anthropic2026claude}, GPT-5.4~\cite{openai2026gpt54}, DeepSeek-V3.2~\cite{deepseek2025v3}, Qwen3-Max~\cite{qwen2025max}, and LLaMA-3.3-70B~\cite{meta2024llama33}. The first four are accessed via their respective vendor APIs, while LLaMA-3.3-70B is deployed locally on our cluster. The local deployment is deliberate: it verifies that SAGE can run entirely within a private network for confidential settings where data forbids the use of public LLM endpoints.

\paragraph{Temperature and randomness}
All LLM calls use temperature $0.1$. The low temperature, combined with the API-level \texttt{seed} parameter described in Section~\ref{sec:metrics}, ensures that each LLM call is reproducible under a fixed seed to the greatest extent possible. The unique residual variance across the five seeds comes from truly distinct train/val/test partitions and the resulting model variations.

\paragraph{Convergence parameters}
We use $K = 20$ (maximum iterations of $\mathcal{A}_3$), $\varepsilon = 0.001$ (reward-improvement tolerance), $k = 4$ (patience window before triggering early stopping), and $\tau_0 = 0.85$ (the default recall-constraint threshold, which $\mathcal{A}_1$ adapts per dataset; e.g., $\tau=0.75$ on IEEE-CIS). The reward weights $(w_1, w_2, w_3, w_4) = (0.40, 0.30, 0.20, 0.10)$ are fixed across all experiments.

\paragraph{Sandbox environment and hardware}
Generated code is executed in an isolated sandbox running Python~3.13.3 with XGBoost~3.1.3, LightGBM~4.6.0, scikit-learn~1.7.2, pandas~2.3.3, NumPy~2.1.3 and other necessary dependencies. All experiments are run on a server with dual AMD EPYC~7543 CPUs (128 cores), 1\,TB RAM, and 8$\times$NVIDIA A40 GPUs (46\,GB each) running Ubuntu 24.04.3 LTS. The LLaMA-3.3-70B backbone is served locally via vLLM on a subset of these GPUs.

\medskip
\noindent
Building on this setup, our evaluation is organized around four research questions (RQs):
\begin{itemize}\setlength\itemsep{2pt}
    \item \textbf{RQ1:} How does SAGE perform on different fraud detection datasets in terms of overall detection performance?
    \item \textbf{RQ2:} How sensitive is SAGE's performance to the choice of the underlying LLM and datasets?
    \item \textbf{RQ3:} How interpretable is SAGE's NLG-guided optimization and reward process?
    \item \textbf{RQ4:} How much does each of the three agents contribute to SAGE's performance?
\end{itemize}

\subsection{Main Results}
\label{sec:main_results}

\begin{figure}[t]
    \centering
    \includegraphics[width=\columnwidth]{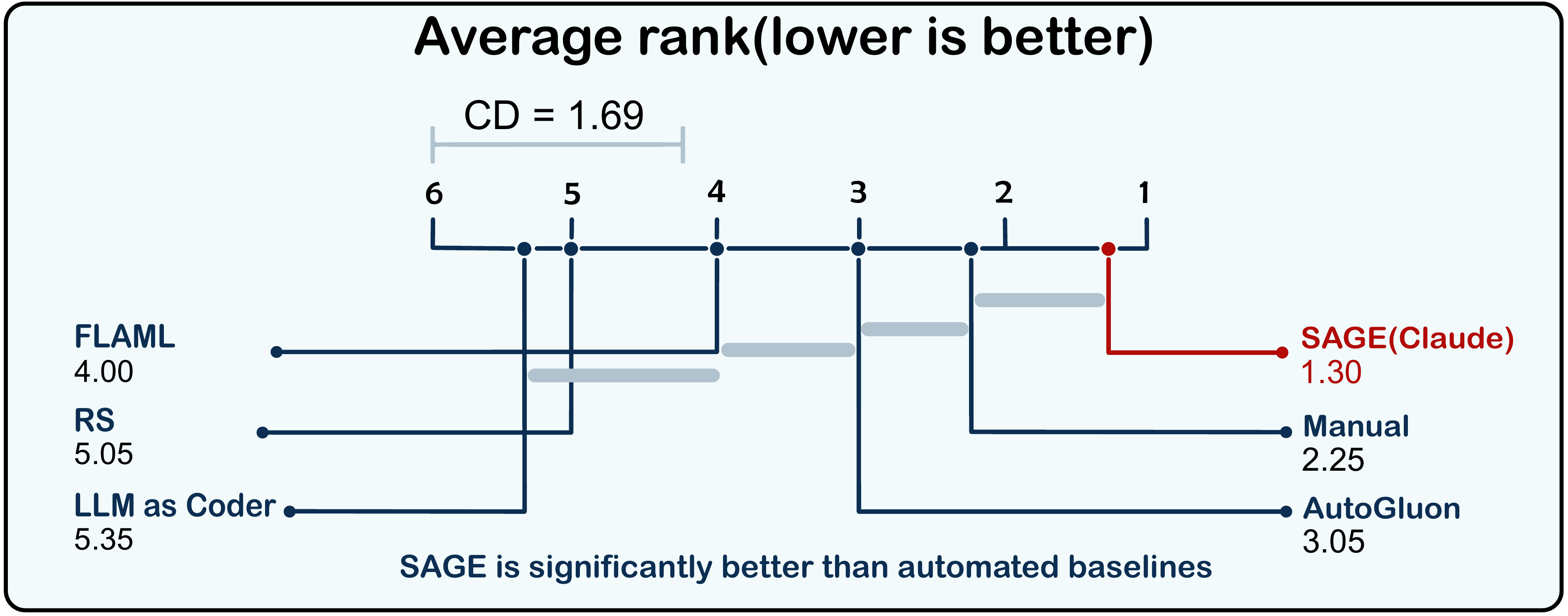}
    \caption{Critical-difference diagram (Friedman + Nemenyi, $\alpha=0.05$) over all $20$ (dataset, metric) blocks.}
    \label{fig:cd_diagram}
\end{figure}

To answer RQ1, we compare SAGE (with Claude Opus~4.7 as the backbone) against five baselines across five datasets and four main metrics. The full results are reported in Table~\ref{tab:main_results}, and we further validate each metric with the Friedman test, complemented by a Friedman--Nemenyi critical-difference analysis (Figure~\ref{fig:cd_diagram}).

\begin{table*}[!t]
    \centering
    \small
    \caption{SAGE versus five baselines on five datasets and four metrics (mean$\pm$std over five seeds). For each (dataset, metric) the \textbf{best} value is in bold and the \underline{second-best} is underlined. For Elliptic and TeleGuard, the test split contains only $\sim$8.4K and $\sim$6.2K negative samples respectively, so the minimum measurable FPR exceeds $10^{-4}$; on these two datasets the reported R@FPR$_{10^{-4}}$ is therefore equivalent to recall at the zero-false-positive operating point.}
    \label{tab:main_results}
    \setlength{\tabcolsep}{4pt}
    \renewcommand{\arraystretch}{0.80}
    \begin{tabularx}{\textwidth}{@{}llCCCC@{}}
        \toprule
        \textbf{Dataset} & \textbf{Method} & \textbf{F1} & \textbf{MCC} & \textbf{R@FPR$_{10^{-4}}$} & \textbf{AUPRC} \\
        \midrule
        \multirow{6}{*}{Credit Card}
          & RS             & 0.8269{\scriptsize$\pm$0.0192} & 0.8278{\scriptsize$\pm$0.0197} & 0.7714{\scriptsize$\pm$0.0522} & 0.8625{\scriptsize$\pm$0.0316} \\
          & Manual         & \underline{0.8577}{\scriptsize$\pm$0.0189} & 0.8582{\scriptsize$\pm$0.0186} & 0.7878{\scriptsize$\pm$0.0497} & 0.8701{\scriptsize$\pm$0.0306} \\
          & FLAML          & 0.6831{\scriptsize$\pm$0.3425} & 0.6852{\scriptsize$\pm$0.3435} & 0.7755{\scriptsize$\pm$0.0540} & 0.8495{\scriptsize$\pm$0.0349} \\
          & AutoGluon      & 0.8563{\scriptsize$\pm$0.0375} & \underline{0.8605}{\scriptsize$\pm$0.0350} & \textbf{0.8041}{\scriptsize$\pm$0.0637} & \underline{0.8717}{\scriptsize$\pm$0.0312} \\
          & LLM as Coder   & 0.8303{\scriptsize$\pm$0.0355} & 0.8354{\scriptsize$\pm$0.0329} & 0.7735{\scriptsize$\pm$0.0773} & 0.8482{\scriptsize$\pm$0.0256} \\
          & \textbf{SAGE}  & \textbf{0.8673}{\scriptsize$\pm$0.0166} & \textbf{0.8684}{\scriptsize$\pm$0.0152} & \underline{0.7918}{\scriptsize$\pm$0.0431} & \textbf{0.8731}{\scriptsize$\pm$0.0281} \\
        \midrule
        \multirow{6}{*}{PaySim}
          & RS             & 0.1593{\scriptsize$\pm$0.0486} & 0.2857{\scriptsize$\pm$0.0506} & 0.0000{\scriptsize$\pm$0.0000} & 0.1166{\scriptsize$\pm$0.0616} \\
          & Manual         & 0.8770{\scriptsize$\pm$0.0074} & 0.8779{\scriptsize$\pm$0.0070} & \underline{0.8847}{\scriptsize$\pm$0.0075} & \underline{0.9601}{\scriptsize$\pm$0.0007} \\
          & FLAML          & \underline{0.8951}{\scriptsize$\pm$0.0209} & \underline{0.8976}{\scriptsize$\pm$0.0193} & 0.8814{\scriptsize$\pm$0.0281} & 0.9503{\scriptsize$\pm$0.0207} \\
          & AutoGluon      & 0.8000{\scriptsize$\pm$0.0349} & 0.8073{\scriptsize$\pm$0.0349} & 0.7156{\scriptsize$\pm$0.0441} & 0.8121{\scriptsize$\pm$0.0273} \\
          & LLM as Coder   & 0.4951{\scriptsize$\pm$0.3162} & 0.5612{\scriptsize$\pm$0.2658} & 0.6875{\scriptsize$\pm$0.3511} & 0.9099{\scriptsize$\pm$0.0643} \\
          & \textbf{SAGE}  & \textbf{0.9973}{\scriptsize$\pm$0.0004} & \textbf{0.9973}{\scriptsize$\pm$0.0004} & \textbf{0.9965}{\scriptsize$\pm$0.0007} & \textbf{0.9976}{\scriptsize$\pm$0.0007} \\
        \midrule
        \multirow{6}{*}{IEEE-CIS}
          & RS             & 0.3804{\scriptsize$\pm$0.0399} & 0.4175{\scriptsize$\pm$0.0363} & 0.1216{\scriptsize$\pm$0.0268} & 0.6522{\scriptsize$\pm$0.0381} \\
          & Manual         & \underline{0.7774}{\scriptsize$\pm$0.0454} & \underline{0.7713}{\scriptsize$\pm$0.0466} & \textbf{0.3799}{\scriptsize$\pm$0.0487} & \textbf{0.8441}{\scriptsize$\pm$0.0175} \\
          & FLAML          & 0.7291{\scriptsize$\pm$0.0674} & 0.7417{\scriptsize$\pm$0.0594} & 0.2899{\scriptsize$\pm$0.1477} & 0.8009{\scriptsize$\pm$0.0641} \\
          & AutoGluon      & 0.6134{\scriptsize$\pm$0.0069} & 0.6444{\scriptsize$\pm$0.0064} & 0.2685{\scriptsize$\pm$0.0182} & 0.6700{\scriptsize$\pm$0.0071} \\
          & LLM as Coder   & 0.3297{\scriptsize$\pm$0.0979} & 0.3555{\scriptsize$\pm$0.0801} & 0.0895{\scriptsize$\pm$0.0530} & 0.5300{\scriptsize$\pm$0.0515} \\
          & \textbf{SAGE}  & \textbf{0.7816}{\scriptsize$\pm$0.0221} & \textbf{0.7752}{\scriptsize$\pm$0.0225} & \underline{0.3781}{\scriptsize$\pm$0.0775} & \underline{0.8338}{\scriptsize$\pm$0.0234} \\
        \midrule
        \multirow{6}{*}{Elliptic}
          & RS             & 0.9496{\scriptsize$\pm$0.0108} & 0.9443{\scriptsize$\pm$0.0120} & 0.7994{\scriptsize$\pm$0.1323} & 0.9866{\scriptsize$\pm$0.0032} \\
          & Manual         & \underline{0.9644}{\scriptsize$\pm$0.0025} & \underline{0.9608}{\scriptsize$\pm$0.0028} & \underline{0.8374}{\scriptsize$\pm$0.0783} & 0.9884{\scriptsize$\pm$0.0019} \\
          & FLAML          & 0.9580{\scriptsize$\pm$0.0035} & 0.9542{\scriptsize$\pm$0.0039} & 0.8007{\scriptsize$\pm$0.1818} & 0.9883{\scriptsize$\pm$0.0031} \\
          & AutoGluon      & 0.9587{\scriptsize$\pm$0.0021} & 0.9552{\scriptsize$\pm$0.0021} & 0.8055{\scriptsize$\pm$0.1788} & \textbf{0.9896}{\scriptsize$\pm$0.0016} \\
          & LLM as Coder   & 0.9324{\scriptsize$\pm$0.0056} & 0.9265{\scriptsize$\pm$0.0068} & 0.8141{\scriptsize$\pm$0.0325} & 0.9788{\scriptsize$\pm$0.0034} \\
          & \textbf{SAGE}  & \textbf{0.9657}{\scriptsize$\pm$0.0014} & \textbf{0.9624}{\scriptsize$\pm$0.0014} & \textbf{0.8592}{\scriptsize$\pm$0.0792} & \underline{0.9894}{\scriptsize$\pm$0.0016} \\
        \midrule
        \multirow{6}{*}{TeleGuard}
          & RS             & 0.9346{\scriptsize$\pm$0.0079} & 0.9342{\scriptsize$\pm$0.0078} & 0.8742{\scriptsize$\pm$0.0413} & 0.9622{\scriptsize$\pm$0.0142} \\
          & Manual         & \underline{0.9513}{\scriptsize$\pm$0.0104} & \underline{0.9509}{\scriptsize$\pm$0.0104} & 0.9161{\scriptsize$\pm$0.0400} & \underline{0.9673}{\scriptsize$\pm$0.0100} \\
          & FLAML          & 0.9295{\scriptsize$\pm$0.0371} & 0.9305{\scriptsize$\pm$0.0358} & 0.7774{\scriptsize$\pm$0.2054} & 0.9584{\scriptsize$\pm$0.0227} \\
          & AutoGluon      & 0.9457{\scriptsize$\pm$0.0203} & 0.9464{\scriptsize$\pm$0.0199} & \underline{0.9194}{\scriptsize$\pm$0.0489} & \textbf{0.9727}{\scriptsize$\pm$0.0110} \\
          & LLM as Coder   & 0.8303{\scriptsize$\pm$0.0363} & 0.8347{\scriptsize$\pm$0.0342} & 0.1677{\scriptsize$\pm$0.2294} & 0.9141{\scriptsize$\pm$0.0304} \\
          & \textbf{SAGE}  & \textbf{0.9545}{\scriptsize$\pm$0.0149} & \textbf{0.9542}{\scriptsize$\pm$0.0151} & \textbf{0.9419}{\scriptsize$\pm$0.0164} & 0.9655{\scriptsize$\pm$0.0084} \\
        \bottomrule
    \end{tabularx}
\end{table*}

\paragraph{Overall performance}
Across the $25$ method--dataset comparisons (five baselines $\times$ five datasets), SAGE wins $24$ (a $96.00\%$ win rate), where a method--dataset pair is counted as a SAGE win if SAGE outperforms the baseline on the majority of the four metrics. SAGE ranks first on $15$ of the $20$ individual (dataset, metric) cells. Its gains are largest on the hardest datasets: on PaySim, whose extreme imbalance (IR$=773.7$) drives most baselines to collapse (RS reaches an F1 of only $0.1593$), SAGE attains $0.9973$, a $135.43\%$ average improvement; on IEEE-CIS, with $433$ heavily missing features, it improves the baseline average by $55.54\%$. On the easier Elliptic and TeleGuard datasets it still secures top or near-top results, showing that its advantage does not sacrifice easy regimes.

\paragraph{Honest analysis}
The few cases where a baseline edges ahead are isolated single-metric wins on otherwise SAGE-dominated datasets, not signs of instability. AutoGluon leads on Credit Card R@FPR$_{10^{-4}}$ ($0.8041$ vs.\ $0.7918$) and on Elliptic/TeleGuard AUPRC, and Manual narrowly leads on IEEE-CIS R@FPR$_{10^{-4}}$ and AUPRC; in every case SAGE wins the remaining metrics on the same dataset. Crucially, SAGE never collapses: FLAML's F1 std reaches $0.3425$ on Credit Card and LLM-as-Coder's R@FPR$_{10^{-4}}$ drops to $0.1677$ on TeleGuard, whereas SAGE's std stays consistently small.

\paragraph{Critical-difference analysis}
We validate significance with the Friedman~\cite{demsar2006statistical} test applied per metric (5 datasets, 6 methods), which rejects equal performance for all four metrics ($p=0.0012$, $0.0015$, $0.0032$, $0.0050$). Pooling all $20$ blocks and adding a Nemenyi post-hoc test yields the CD diagram in Figure~\ref{fig:cd_diagram}: SAGE attains the best average rank ($1.30$), ahead of Manual ($2.25$), AutoGluon ($3.05$), FLAML ($4.00$), RS ($5.05$), and LLM as Coder ($5.35$), with the pooled test highly significant ($\chi^2=72.46$, $p<10^{-13}$). At $\mathrm{CD}=1.69$, SAGE is significantly better than all baselines except Manual, whose gap ($0.95$) falls below the CD.

\begin{tcolorbox}[
    enhanced,
    colback=gray!8,
    colframe=gray!50,
    boxrule=0pt,
    arc=1mm,
    left=8pt, right=8pt, top=8pt, bottom=8pt,
    title=\textbf{Takeaway1},
    fonttitle=\bfseries\color{white},
    colbacktitle=gray!55,
    coltitle=white,
    attach boxed title to top left={xshift=0mm, yshift=0mm},
    boxed title style={colframe=gray!55, sharp corners, boxrule=0pt}
]
SAGE wins $96.00\%$ of the $25$ method--dataset comparisons and attains the best average rank ($1.30$), with statistically significant superiority over all automated baselines. It reaches, and on most metrics surpasses, expert human data scientists without any manual intervention.
\end{tcolorbox}

\subsection{Robustness and Sensitivity}
\label{sec:robustness}

To answer RQ2, we examine whether the effectiveness of SAGE depends on the specific LLM driving its agents. We re-run the entire pipeline on all five datasets using five backbone: Claude Opus~4.7, GPT-5.4, DeepSeek-V3.2, Qwen3-Max, and LLaMA-3.3-70B, and analyze both the win rate against the baselines and the performance spread across backbones.

\begin{figure}[H]
    \centering
    \includegraphics[width=0.96\columnwidth]{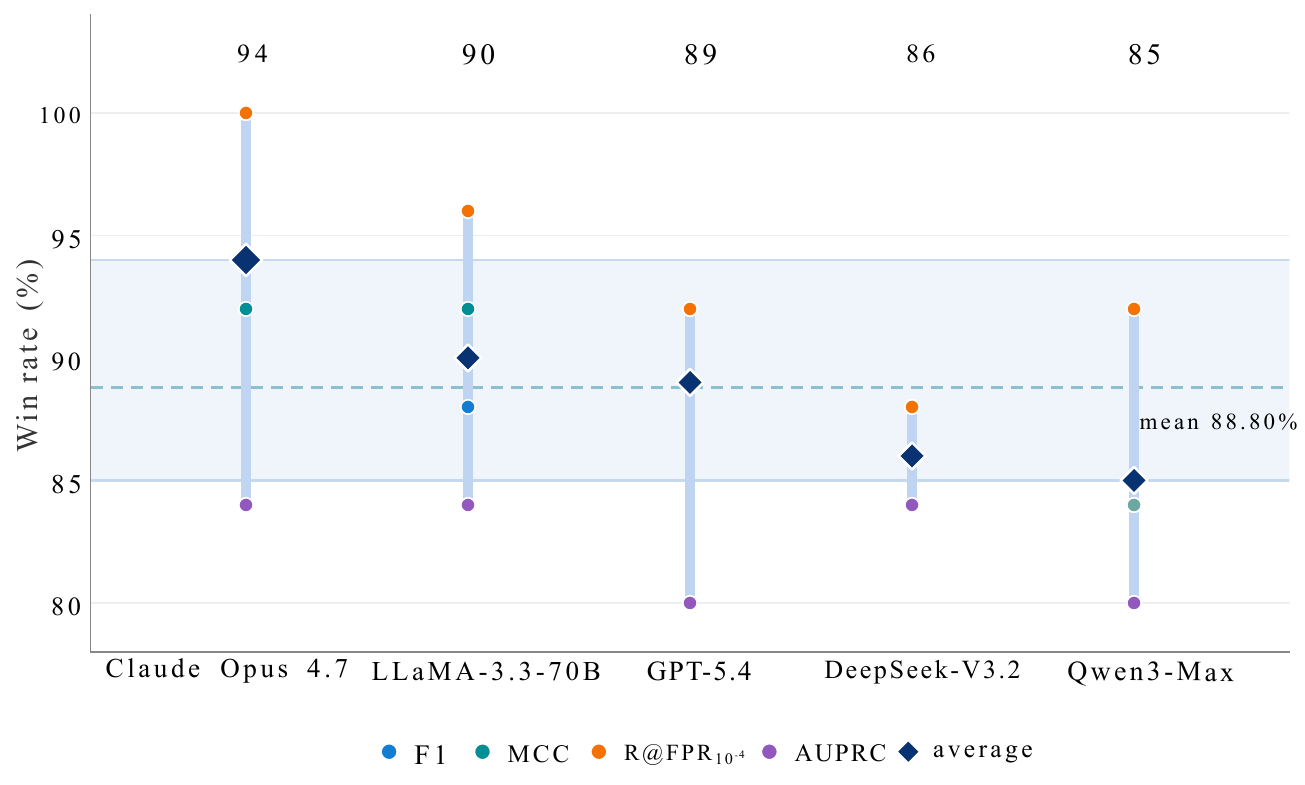}
    \caption{Per-metric win rate of SAGE over the $25$ method--dataset comparisons (five baselines $\times$ five datasets) under five LLM backbones, with the four-metric average shown on the right.}
    \label{fig:backbone_winrate}
\end{figure}

\paragraph{Cross-backbone Robustness}
Figure~\ref{fig:backbone_winrate} reports, for each backbone, the win rate of SAGE over the $25$ method--dataset comparisons (five baselines $\times$ five datasets). All five backbones place SAGE clearly ahead of the baseline field: the win rate ranges from $94.00\%$ (Claude Opus~4.7) down to $85.00\%$ (Qwen3-Max), a spread of only $9.00$ percentage points, with an average of $88.80\%$. Even the weakest backbone wins more than four out of five comparisons, and the locally deployed LLaMA-3.3-70B reaches $90.00\%$, confirming that SAGE delivers strong results even under the on-premise, privacy-preserving deployment required for confidential data Figure~\ref{fig:backbone_winrate} uses a finer granularity than the $96.00\%$ figure in Section~\ref{sec:main_results}: the latter counts a win at the method--dataset pair level, whereas Figure~\ref{fig:backbone_winrate} counts wins independently on each metric, so Claude Opus 4.7's per-metric average ($94.00\%$) is consistent with, and slightly below, the aggregate $96.00\%$. A complementary perspective is the \emph{spread}: a smaller spread on the main metrics means the framework is less sensitive to the choice of backbone. As summarized in Table~\ref{tab:backbone_spread}, SAGE's average AUPRC spread is only $1.13\%$ and its average F1 spread is only $3.60\%$.

\begin{table}[!t]
    \centering
    \caption{Cross-backbone performance spread per dataset, reported as the relative range across the five LLM backbones. The last row gives the average over datasets.}
    \label{tab:backbone_spread}
    \scriptsize
    \renewcommand{\arraystretch}{1.0}
    \setlength{\tabcolsep}{6pt}
    \begin{tabular}{lcccc}
        \toprule
        \textbf{Dataset} & \textbf{AUPRC Spread} & \textbf{AUPRC Range} & \textbf{F1 Spread} & \textbf{F1 Range} \\
        \midrule
        Credit Card & 0.0034 & 0.39\%  & 0.0076 & 0.87\%  \\
        PaySim      & 0.0028 & 0.28\%  & 0.0185 & 1.85\%  \\
        IEEE-CIS    & 0.0392 & 4.70\%  & 0.1004 & 12.85\% \\
        Elliptic    & 0.0007 & 0.07\%  & 0.0029 & 0.30\%  \\
        TeleGuard   & 0.0022 & 0.23\%  & 0.0206 & 2.13\%  \\
        \midrule
        \textbf{Average} & --- & \textbf{1.13\%} & --- & \textbf{3.60\%} \\
        \bottomrule
    \end{tabular}
\end{table}

\begin{figure}[ht]
    \centering
    \includegraphics[width=\columnwidth]{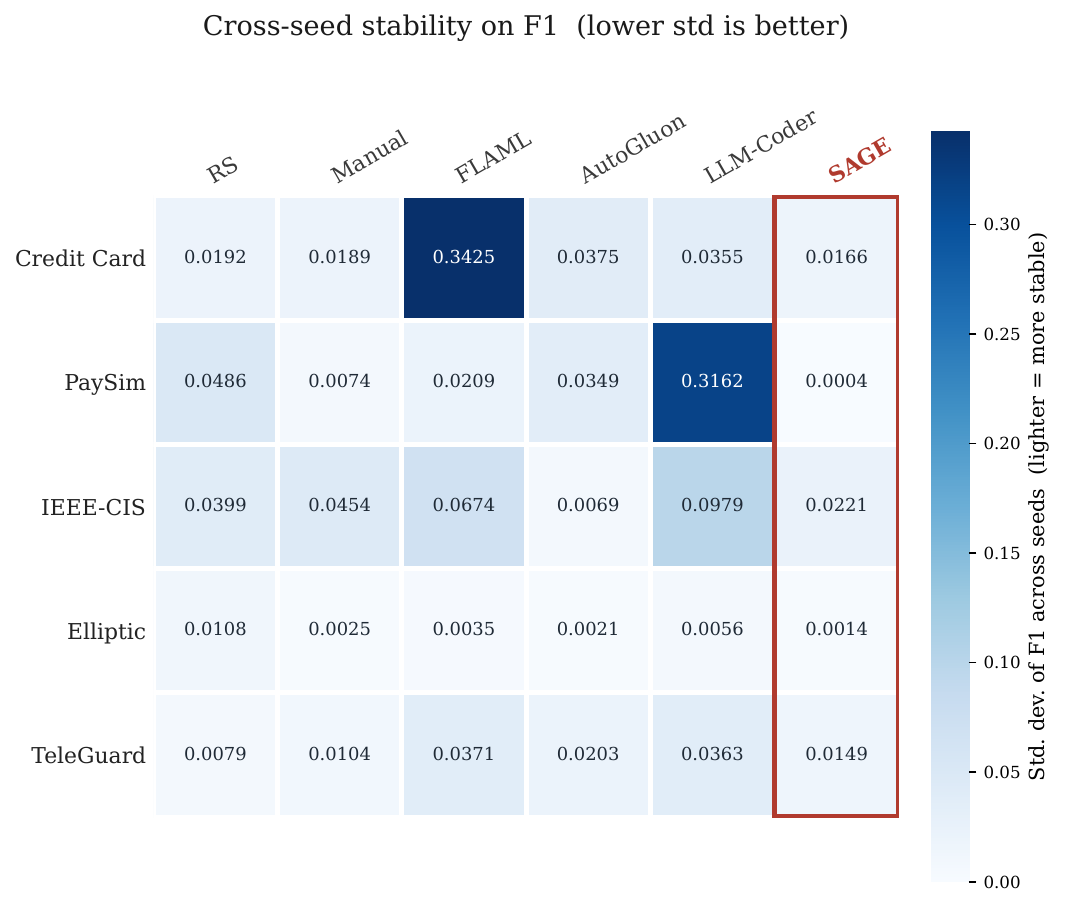}
    \caption{Per-seed standard deviation across the five datasets (lighter = more stable). SAGE attains the lowest or near-lowest F1 standard deviation on every dataset.}
    \label{fig:std_heatmap}
\end{figure}

\paragraph{Sensitivity across data splits}
Besides its robustness to the backbone, SAGE is also the least sensitive to the five random data splits. Returning to the main results (Table~\ref{tab:main_results}), SAGE achieves the smallest standard deviation on the vast majority of (dataset, metric) cells, whereas the baseline methods are highly sensitive in the worst cases: FLAML's F1 standard deviation on Credit Card reaches $0.3425$, as a single failed split causes its mean to plummet, while the LLM-as-Coder baseline fluctuates by $0.3162$ in F1 on PaySim and drops to $0.1677$ in R@FPR$_{10^{-4}}$ on TeleGuard. In contrast, SAGE's standard deviation remains consistently small: for example, $0.0004$ in F1 on PaySim and only $0.0166$ in F1 on Credit Card. Figure~\ref{fig:std_heatmap} visualizes this contrast as a per-seed standard-deviation heatmap on the F1 metric, where SAGE is consistently among the most stable methods, attaining the lowest average F1 standard deviation on the five datasets. Crucially, unlike FLAML and LLM-as-Coder, SAGE never exhibits a catastrophic worst case.

\begin{tcolorbox}[
    enhanced,
    colback=gray!8,
    colframe=gray!50,
    boxrule=0pt,
    arc=1mm,
    left=8pt, right=8pt, top=8pt, bottom=8pt,
    title=\textbf{Takeaway2},
    fonttitle=\bfseries\color{white},
    colbacktitle=gray!55,
    coltitle=white,
    attach boxed title to top left={xshift=0mm, yshift=0mm},
    boxed title style={colframe=gray!55, sharp corners, boxrule=0pt}
]
SAGE's performance is largely insensitive to the underlying LLM: all five backbones keep its win rate above $85\%$ (average $88.80\%$), with an average spread of only $1.13\%$ in AUPRC and $3.60\%$ in F1 across datasets. It is also the least sensitive method to the random data splits, showing that the framework, rather than any specific model or partition, drives its effectiveness.
\end{tcolorbox}

\subsection{Case Study}
\label{sec:case_study}

\begin{figure}[ht]
    \centering
    \includegraphics[width=\columnwidth]{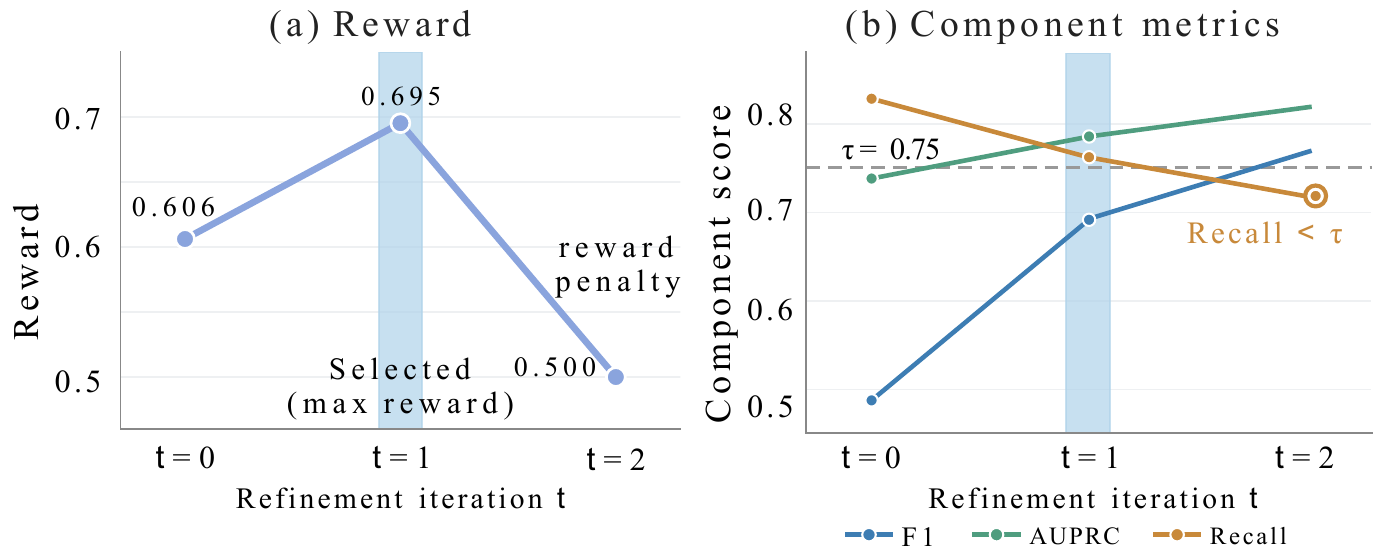}
    \caption{Reward and underlying metrics over optimization iterations on IEEE-CIS.}
    \label{fig:reward_curve}
\end{figure}

To answer RQ3, we examine the interpretability of SAGE's NLG-guided optimization and reward process. We trace a single run on IEEE-CIS, using Claude Opus~4.7 as the backbone, and examine the critiques generated by $\mathcal{A}_3$ and their reward trajectory. Figure~\ref{fig:reward_curve} shows the reward and related metrics as a function of the number of optimization iterations.

\paragraph{Interpretable Diagnosis and Action}
At each iteration, the critic LLM emits a structured natural-language gradient organized as \textsc{Observation}, \textsc{Diagnosis}, and \textsc{Action}, which is both human-readable and directly auditable. The box below reproduces the complete critique for the first iteration. This diagnosis is not generic but is generated by the LLM from the characteristics of each iteration: it correctly attributes the initial model's precision collapse ($0.3447$) to the interaction between an aggressive \texttt{scale\_pos\_weight}$=27.58$ and the default decision threshold of $0.50$, and further points out that the high-signal IEEE-fraud columns (\texttt{TransactionAmt}, the C/D groups, \texttt{card1}, \texttt{addr1}) are left unexploited. The corresponding actions are concrete and traceable: raising the threshold, reducing the positive weight, deepening the model, and adding domain-specific engineered features, each mapped to a specific optimization handle in the generated code, turning the LLM reasoning into a verifiable, reproducible edit history.


\paragraph{Reward-Based Multi-Objective Trade-offs}
Figure~\ref{fig:reward_curve} shows that the optimization is not simply about increasing the F1 score, but jointly weighs detection quality and operational constraints at every iteration. Iteration $t=1$ increases the F1 score from $0.4866$ to $0.6907$ while maintaining recall above the threshold $\tau$, and restores precision from $0.3447$ to $0.6319$, thus achieving a peak reward of $0.6954$. At $t=2$, the agent further advances feature engineering, obtaining a higher F1 score ($0.7658$) and AUPRC ($0.8177$), but its recall drops to $0.7176$, below $\tau$, triggering the recall-constraint gate $W_{\mathrm{recall}}$ in the composite reward and causing the reward to drop to $0.5001$. Because SAGE selects the iteration that maximizes the reward, i.e., $t^{*}=\arg\max_t R(\mathbf{m}_t)$, it correctly preserves the recall-compliant $t=1$ model rather than the $t=2$ model that superficially has a higher F1 score. Therefore, within the SAGE framework, a higher F1 score obtained by sacrificing the recall constraint is rejected, indicating that the reward performs a true multi-objective trade-off rather than pursuing a single metric, which is the exact behavior production fraud systems require when scoring candidate models against business-defined risk budgets.

\begin{tcolorbox}[
    enhanced, unbreakable,
    colback=white,
    colframe=gray!50,
    boxrule=0.6pt,
    arc=1mm,
    left=8pt, right=8pt, top=8pt, bottom=8pt,
    fonttitle=\bfseries\small\color{white},
    colbacktitle=gray!70,
    coltitle=white,
    attach boxed title to top left={xshift=8pt, yshift=-2.5pt},
    boxed title style={colframe=gray!70, sharp corners, arc=0.5mm, boxrule=0pt},
    title={NLG critique $D_1$ at iteration $t=1$ on IEEE-CIS},
    label={box:nlg_example}
]

\begin{tcolorbox}[
    enhanced, breakable, nobeforeafter,
    colback=blue!18, colframe=blue!70!black,
    boxrule=0.8pt, arc=1mm,
    left=6pt, right=6pt, top=4pt, bottom=4pt,
    fontupper=\footnotesize\ttfamily,
    title=\textsc{Observation}, fonttitle=\bfseries\footnotesize\color{white},
    colbacktitle=blue!70!black, coltitle=white,
    attach boxed title to top left={xshift=6pt, yshift=-2pt},
    boxed title style={colframe=blue!70!black, sharp corners, boxrule=0pt}
]
AUPRC=0.7373 ($<$tau=0.75), F1=0.4866 low due to Precision=0.3447, Recall=0.8275 strong.
\end{tcolorbox}

\vspace{5pt}

\begin{tcolorbox}[
    enhanced, breakable, nobeforeafter,
    colback=orange!22, colframe=orange!85!black,
    boxrule=0.8pt, arc=1mm,
    left=6pt, right=6pt, top=4pt, bottom=4pt,
    fontupper=\footnotesize\ttfamily,
    title=\textsc{Diagnosis}, fonttitle=\bfseries\footnotesize\color{white},
    colbacktitle=orange!85!black, coltitle=white,
    attach boxed title to top left={xshift=6pt, yshift=-2pt},
    boxed title style={colframe=orange!85!black, sharp corners, boxrule=0pt}
]
High scale\_pos\_weight=27.58 combined with threshold=0.50 inflates positive predictions causing precision collapse; also missing engineered features on rich IEEE-fraud columns (TransactionAmt, C1/C13/C14, D1/D2/D15, card1/addr1/emaildomain) which are known high-signal.
\end{tcolorbox}

\vspace{5pt}

\begin{tcolorbox}[
    enhanced, breakable, nobeforeafter,
    colback=green!20, colframe=green!55!black,
    boxrule=0.8pt, arc=1mm,
    left=6pt, right=6pt, top=4pt, bottom=4pt,
    fontupper=\footnotesize\ttfamily,
    title=\textsc{Action}, fonttitle=\bfseries\footnotesize\color{white},
    colbacktitle=green!55!black, coltitle=white,
    attach boxed title to top left={xshift=6pt, yshift=-2pt},
    boxed title style={colframe=green!55!black, sharp corners, boxrule=0pt}
]
(1) Raise threshold 0.50$\rightarrow$0.55 to restore precision. (2) Reduce scale\_pos\_weight 27.58$\rightarrow$15.0 to reduce false positives while keeping recall. (3) Strengthen model: n\_estimators 500$\rightarrow$800, max\_depth 7$\rightarrow$8, learning\_rate 0.05$\rightarrow$0.03. (4) Add features: log1p(TransactionAmt), C1/C13 and C14/C1 ratios, D1$-$D2, D15/D1, frequency encoding of card1/addr1/P\_emaildomain, card1$\times$card2.
\end{tcolorbox}
\end{tcolorbox}

\begin{tcolorbox}[
    enhanced,
    colback=gray!8,
    colframe=gray!50,
    boxrule=0pt,
    arc=1mm,
    left=8pt, right=8pt, top=8pt, bottom=8pt,
    title=\textbf{Takeaway3},
    fonttitle=\bfseries\color{white},
    colbacktitle=gray!55,
    coltitle=white,
    attach boxed title to top left={xshift=0mm, yshift=0mm},
    boxed title style={colframe=gray!55, sharp corners, boxrule=0pt}
]
SAGE's optimization is fully interpretable: each iteration emits a human-readable Observation--Diagnosis--Action critique that pinpoints concrete failure causes and maps them to specific code edits. Its reward process is equally transparent, performing a genuine multi-objective trade-off that rejects a higher-F1 model violating the recall constraint in favor of the constraint-satisfying one.
\end{tcolorbox}

\subsection{Ablation Study}
\label{sec:ablation}

To answer RQ4, we conduct an ablation study on the IEEE-CIS benchmark to quantify the contribution of the three agents, using Claude Opus~4.7 as the backbone. Starting from the complete SAGE pipeline, we disable one agent at a time and measure the resulting performance change, reported as the mean$\pm$std of F1 over five seeds. Removing $\mathcal{A}_1$ withholds the DDT and the other data-aware structures of the Profiling Agent, so the downstream agents run without data-aware guidance. Removing $\mathcal{A}_2$ replaces its DDT-driven algorithm selection with a fixed Random Forest model as the initial code template, disabling adaptive initial code selection and generation. Removing $\mathcal{A}_3$ skips the NLG-guided optimization loop, so the evaluation reflects only the initial code produced by $\mathcal{A}_2$. Table~\ref{tab:ablation} reports the results.

\paragraph{Each agent contributes}
Removing any one agent lowers the F1 score, confirming that all three agents are necessary. $\mathcal{A}_3$ is the most critical: disabling it causes a $36.19\%$ drop in F1, demonstrating that data analysis and algorithm selection alone cannot produce a high-quality detector. $\mathcal{A}_2$ follows closely: replacing its LLM-based algorithm selection with a fixed Random Forest code template causes a $25.74\%$ drop in F1. Even a fully executed optimization loop by $\mathcal{A}_3$ does not fully recover from a poor initial algorithm choice on IEEE-CIS, suggesting that the model-selection reasoning of $\mathcal{A}_2$ is a key contributor. $\mathcal{A}_1$ provides essential guidance: removing its DDT and semantic interpretation causes a $13.72\%$ drop in F1, leaving the downstream agents with only generic strategies and lacking a holistic understanding of the dataset.


\begin{table}[!t]
    \centering
    \scriptsize
    \caption{Ablation on IEEE-CIS. A \textcolor{green!55!black}{\checkmark} indicates the agent is enabled and a \textcolor{red!75!black}{\ding{55}} indicates it is removed. $\Delta$F1 is the relative F1 drop with respect to the full SAGE pipeline.}
    \label{tab:ablation}
    \renewcommand{\arraystretch}{1.15}
    \resizebox{\linewidth}{!}{
    \begin{tabular}{lccccc}
        \toprule
        \textbf{Configuration} & $\bm{\mathcal{A}_1}$ & $\bm{\mathcal{A}_2}$ & $\bm{\mathcal{A}_3}$ & \textbf{F1} & \textbf{$\Delta$F1} \\
        \midrule
        w/o $\mathcal{A}_1$
        & \textcolor{red!75!black}{\ding{55}}
        & \textcolor{green!55!black}{\checkmark}
        & \textcolor{green!55!black}{\checkmark}
        & 0.6744{\scriptsize$\,\pm\,$0.0307}
        & $-13.72\%$ \\

        w/o $\mathcal{A}_2$
        & \textcolor{green!55!black}{\checkmark}
        & \textcolor{red!75!black}{\ding{55}}
        & \textcolor{green!55!black}{\checkmark}
        & 0.5804{\scriptsize$\,\pm\,$0.0445}
        & $-25.74\%$ \\

        w/o $\mathcal{A}_3$
        & \textcolor{green!55!black}{\checkmark}
        & \textcolor{green!55!black}{\checkmark}
        & \textcolor{red!75!black}{\ding{55}}
        & 0.4988{\scriptsize$\,\pm\,$0.0391}
        & $-36.19\%$ \\
        \midrule
        \textbf{Full SAGE}
        & \textcolor{green!55!black}{\checkmark}
        & \textcolor{green!55!black}{\checkmark}
        & \textcolor{green!55!black}{\checkmark}
        & \textbf{0.7816}{\scriptsize$\,\pm\,$0.0221}
        & \textbf{---} \\
        \bottomrule
    \end{tabular}
    }
\end{table}

\begin{tcolorbox}[
    enhanced,
    colback=gray!8,
    colframe=gray!50,
    boxrule=0pt,
    arc=1mm,
    left=8pt, right=8pt, top=8pt, bottom=8pt,
    title=\textbf{Takeaway4},
    fonttitle=\bfseries\color{white},
    colbacktitle=gray!55,
    coltitle=white,
    attach boxed title to top left={xshift=0mm, yshift=0mm},
    boxed title style={colframe=gray!55, sharp corners, boxrule=0pt}
]
The three agents form a progressive value chain: $\mathcal{A}_1$ supplies data insight, $\mathcal{A}_2$ makes the optimal algorithmic decision, and $\mathcal{A}_3$ pushes performance to its ceiling through iterative optimization. Their contributions are complementary rather than redundant, and removing any one degrades performance.
\end{tcolorbox}

%% file: sections/06_discussion.tex
\section{Discussion}
\label{sec:discussion}

Our discussion reveals some limitations of SAGE, which also point to specific directions for future research. First, as observed in the honest analysis (Section~\ref{sec:main_results}), SAGE currently generates a single classification model rather than a multi-model ensemble model. On low-dimensional and well-behaved datasets such as Credit Card, AutoGluon's re-stacking method achieves a slightly higher R@FPR value; on Elliptic and TeleGuard, its ensemble model also maintains a slight advantage in the AUPRC metric. Therefore, one optimization direction is to allow the agent to coordinate the stacking layers in the code space, attempting to combine multiple model variants under a guided reward mechanism, so that SAGE can inherit the advantages of a single method without sacrificing interpretability. Recent research on hybrid architectures~\cite{chen2025tlk} has demonstrated that a Transformer-LSTM-KELM scheme utilizing the synergistic complementarity of three heterogeneous modules outperforms nine mainstream baseline classifiers, including XGBoost and CatBoost. Second, SAGE's current evaluation is limited to individual-level tabular fraud and does not leverage the relational signals available in some real settings. Integrating graph-derived features as an additional view in the DDT is therefore a natural next step toward a more relationally aware agentic anti-fraud framework.

%% file: sections/07_conclusion.tex
\section{Conclusion}
\label{sec:conclusion}

This paper proposes a novel multi-agent framework called SAGE for detecting fraudulent behavior in tabular data at the individual level. Its main contribution lies in addressing the under-explored intersection between LLM-driven agent reasoning and the practical constraints of production-grade anti-fraud systems. Traditional AutoML, graph-based methods, and general LLM agents all suffer from shortcomings in four aspects: agent-centricity, tabular data processing capabilities, interpretability, and fraud detection specificity. To our knowledge, SAGE is the first end-to-end LLM-driven agent framework specifically built for tabular fraud detection. It coordinates data understanding, modeling, and reflective optimization through three dedicated agents. Our research demonstrates that building the process on structured data diagnostic priors and setting the optimization process as a reward-driven search guided by natural language comments can transform the previously open agent loop into a process that meets the constraints of anti-fraud business requirements. We assess SAGE on multiple authoritative fraud datasets and across multiple LLM backbones in order to distinguish the contributions of the framework itself from those of the underlying language models. Empirical results confirm that SAGE outperforms strong AutoML, human-expert, and LLM-as-coder baselines.